\documentclass[iicol]{sn-jnl}% Default with double column layout
\jyear{2021}%

\theoremstyle{thmstyleone}%
\theoremstyle{thmstyletwo}%

\theoremstyle{thmstylethree}%

\newcommand{\ie}{\textit{i}.\textit{e}.}
\newcommand{\eg}{\textit{e}.\textit{g}.}
\newcommand{\etal}{\textit{et al}.}

\usepackage{graphicx}
\usepackage{amsmath}
\usepackage{amssymb}
\usepackage{booktabs}
\usepackage{comment}
\usepackage{multirow}
\usepackage{mathtools}
\usepackage{epsfig}
\usepackage[misc]{ifsym}

\usepackage{makecell}
\usepackage{pifont}
\usepackage{bbding}
\usepackage{wrapfig}
\usepackage{verbatim}
\usepackage{relsize}
\usepackage{colortbl}
\usepackage{bm}
\usepackage{bbm}
\usepackage{breakurl}
\usepackage{color}
\usepackage{diagbox}
\usepackage[capitalize]{cleveref}
\usepackage{fancyvrb,cprotect}

\begin{document}

\title[Article Title]{Generalized Few-Shot Continual Learning with Contrastive Mixture of Adapters}


\author[1]{\fnm{Yawen} \sur{Cui}}\email{yawen.cui@oulu.fi}

\author[2]{\fnm{Zitong} \sur{Yu}}\email{zitong.yu@ieee.org}

\author[2]{\fnm{Rizhao} \sur{Cai}}\email{rzcai@ntu.edu.sg}

\author[3]{\fnm{Xun} \sur{Wang}}\email{bnuwangxun@gmail.com}

\author[2]{\fnm{Alex C.} \sur{Kot}}\email{eackot@ntu.edu.sg}

\author[4,1]{\fnm{Li} \sur{Liu\Letter}}\email{dreamliu2010@gmail.com}

%\affil[1]{\orgdiv{School of Electrical and Electronic Engineering}, \orgname{Nanyang Technological University}, \orgaddress{\postcode{639798}, \country{Singapore}}}

\affil[1]{\orgdiv{Center for Machine Vision and Signal Analysis}, \orgname{University of Oulu}, \orgaddress{\city{Oulu}, \country{Finland}}}

\affil[2]{\orgdiv{School of Electrical and Electronic Engineering}, \orgname{Nanyang Technological University}}

\affil[3]{\orgdiv{ByteDance}, \orgaddress{\city{Beijing}, \country{China}}}

\affil[4]{\orgdiv{College of Electronic Science}, \orgname{National University of Defense Technology}, 
\orgaddress{\city{Changsha}, \country{China}}}

%\affil[2]{\orgdiv{Department of Engineering}, \orgname{University of Oxford}, \orgaddress{\city{Oxford}, \postcode{OX13PJ}, \country{UK}}}

%\affil[2]{\orgdiv{School of Software Engineering}, \orgname{Xi'an Jiaotong University}, \orgaddress{ \city{Xi'an}, \postcode{710049}, \country{China}}}


\abstract{The goal of Few-Shot Continual Learning (FSCL) is to incrementally learn novel tasks with limited labeled samples and preserve previous capabilities simultaneously, while current FSCL methods are all for the class-incremental purpose. Moreover, the evaluation of FSCL solutions is only the cumulative performance of all encountered tasks, but there is no work on exploring the domain generalization ability. Domain generalization is a challenging yet practical task that aims to generalize beyond training domains. In this paper, we set up a Generalized FSCL (GFSCL) protocol involving both class- and domain-incremental situations together with the domain generalization assessment. Firstly, two benchmark datasets and protocols are newly arranged, and detailed baselines are provided for this unexplored configuration. We find that common continual learning methods have poor generalization ability on unseen domains and cannot better cope with the catastrophic forgetting issue in cross-incremental tasks. In this way, we further propose a rehearsal-free framework based on Vision Transformer (ViT) named Contrastive Mixture of Adapters (CMoA). Due to different optimization targets of class increment and domain increment, the CMoA contains two parts: (1) For the class-incremental issue, the Mixture of Adapters (MoA) module is incorporated into ViT, then cosine similarity regularization and the dynamic weighting are designed to make each adapter learn specific knowledge and concentrate on particular classes. (2) For the domain-related issues and domain-invariant representation learning, we alleviate the inner-class variation by prototype-calibrated contrastive learning. Finally, six evaluation indicators are compared by comprehensive experiments on the two benchmark datasets to validate the efficacy of CMoA, and the results illustrate that CMoA can achieve comparative performance with rehearsal-based continual learning methods. The codes and protocols are available at \href{https://github.com/yawencui/CMoA}{https://github.com/yawencui/CMoA}. }




\maketitle

\section{Introduction}\label{sec1}

Generally, the supervised learning model is only modified with independent and identically distributed (i.i.d.) data, and concentrates on the specific task. On the contrary, the target of Continual Learning (CL)~\cite{de2021continual,wang2022learning, douillard2022dytox, yan2022learning, xue2022meta, simon2022generalizing} is to learn a sequence of tasks with a single model and alleviate the performance deterioration of previously seen tasks. There are mainly three configurations of CL: (1) Task-incremental learning~\cite{tang2020graph} aims at incrementally learning a sequence of disjoint tasks, which requires the task identity in the prediction procedure. (2) Class-incremental learning (CIL)~\cite{hou2019learning, rebuffi2017icarl} is to construct a unified classifier for all encountered classes at different stages. (3) Domain-incremental learning~\cite{wang2022s, mirza2022efficient} targets to progressively learn categories in novel domains. Several works~\cite{wang2022learning, kim2021cored, fini2022self, marra2019incremental,khan2021video, shon2022dlcft} propose the general approach that can deal with the class-incremental or domain-incremental task, or directly applied class-incremental methods to domain-incremental tasks. To cope with real-world scenarios, Xie~\textit{et al.}~\cite{xie2022general} propose a new configuration by introducing both the class-incremental and domain-incremental tasks into a learning sequence. Simon~\textit{et al.}~\cite{simon2022generalizing} also tackles the CL problem with this two-level increment, while it evaluates the generalization performance on unseen domains, which is different from vanilla CL that evaluates the overall performance on all seen tasks. 

Few-Shot Learning (FSL)~\cite{wang2020generalizing} is a surging topic that deals with the situation that only limited labeled samples are provided. Few-Shot Continual Learning (FSCL)~\cite{tao2020few, zhang2021few, cheraghian2021synthesized, dong2021few, cheraghian2021semantic, zhu2021self, shi2021overcoming, zhou2022forward, ahmad2022few, hersche2022constrained, zhou2022few, chi2022metafscil, peng2022few} is proposed to handle the CL scenario when novel tasks are all FSL tasks. The goal of FSCL is to learn novel tasks with limited labeled samples cumulatively and maintain previous capabilities simultaneously. However, all the existing FSCL works focus on the class-incremental task, which can not handle the practical scenarios of encountered novel domains, such as blurring, fogging, rotation, and scaling. In this paper, we consider this ignored scenario and put forward a more generic configuration considering the class- and domain-incremental tasks together with the performance evaluation on unseen domains, which is Generalized Few-Shot Continual Learning (GFSCL).

\begin{figure*}[t]
\includegraphics[width=0.8\linewidth]{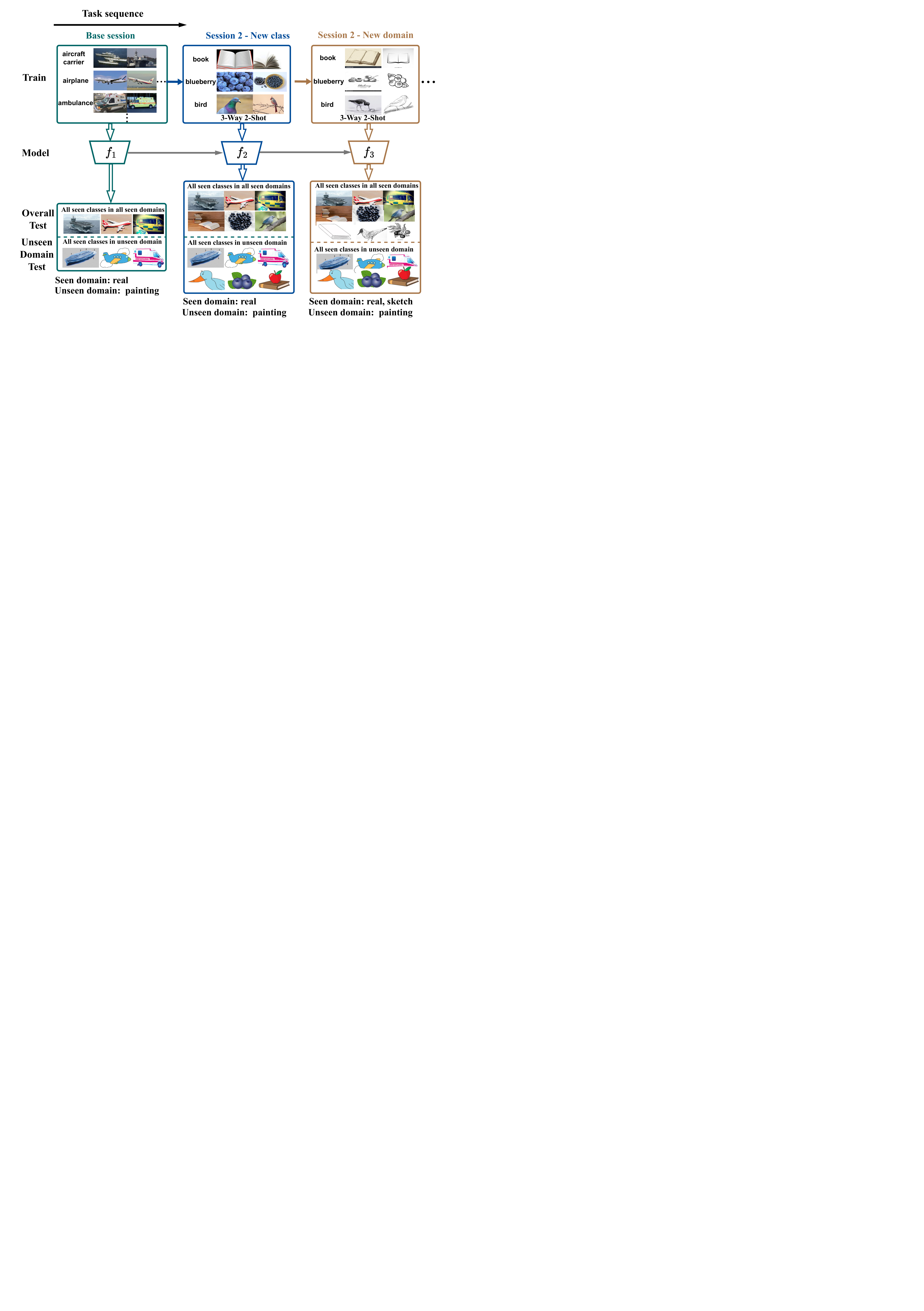}
\centering
\caption{GFSCL configuration. Given a task sequence, the base task is featured by a large dataset containing several categories belonging to one or multiple domains. In following incremental sessions, each task is class or domain-incremental setting and only contains limited training samples (\eg, 3-way 2-shot). During testing, we evaluate the overall accuracy of all seen categories in all seen domains and the generalization ability in unseen domains.}
\label{configuration} 
\end{figure*}

The configuration of GFSCL is illustrated in Fig.~\ref{configuration}: For the task sequence, the base task is featured by a large-scale dataset containing categories belonging to one or multiple domains. In incremental sessions, each task is class-incremental or domain-incremental setting and only contains limited training samples. Moreover, class increment and domain increment are encountered alternatively, which is more challenging than completing one increment first~\cite{xie2022general}. For the testing procedure, we evaluate the overall accuracy of all seen categories in all seen domains. Moreover, we evaluate the generalization ability of all seen categories in unseen domains, which is never considered in previous FSCL methods. There are three main challenges: (1) Fulfilling the stability-plasticity dilemma, which is both stable to prevent forgetting on seen classes and domains as well as plastic to learn new ones. (2) Mitigating the overfitting issue on novel categories or old categories from new domains arrived in incremental sessions. (3) Improving the generalization ability in unseen domains. It is worth noting that the optimization targets of class increment and domain increment are opposite. For class-incremental sessions, the newly-encountered categories should be far from all previous categories to increase inter-class separability. However, for the domain-incremental sessions, current samples of the new domain should be close to the previously-seen samples of the same class to decrease intra-class variation. 

In order to benchmark our proposed GFSCL setting, two datasets and detailed protocols are built first delicately. Based on three widely-used frameworks and popular CL methods, we provide extensive baselines of the GFSCL for the following comparison. When executing GFSCL with the current continual learning methods, we find that not only the performance of domain generalization ability is poor, but also that there is no significant performance improvement in the domain-incremental session. To better solve the GFSCL and protect data privacy, we propose a rehearsal-free framework based on Vision Transformer (ViT) named Contrastive Mixture of Adapters (CMoA). Due to the opposite optimization targets of class-incremental tasks and domain-incremental tasks, we separately premeditate the GFSCL task and propose two parts for the class-incremental task and domain-related issues (\ie, domain-incremental task and domain generalization), respectively.

Incorporating Adapters ~\cite{houlsby2019parameter, jie2022convolutional} into each transformer layer is one of the parameter-efficient tuning methods for ViT, which is suitable for the fast adaptation with limited labeled samples~\cite{bansal2022meta, li2022cross, zhang2022tip}. However, the performance of the previous task is difficult to maintain when the model adapts fast to novel tasks. Therefore, 
we incorporate the mixture of adapters (MoA) module into each transformer layer to solve the class-incremental task in GFSCL, and we hope that each adapter can focus on specific classes for stability purpose. To achieve this aim, the dynamic weighting strategy and the cosine similarity regularization are designed to force each adapter to learn particular knowledge.

%Though current continual learning methods can tackle class-incremental or domain-incremental learning separately, they never consider promoting the domain generalization ability of the continual learning model.

\begin{table*}[t]
\caption{The brief summary of the related research topics.}
%Our proposed UaD-CE achieves the state-of-the-art results with respect to the three indicators.
\centering
\setlength{\tabcolsep}{11pt}
\resizebox{1.0\textwidth}{!}{
\tabcolsep 0.06in
\begin{tabular}{c|c|c|c|c|c|c|c}
\bottomrule[1.3pt]
{\diagbox{\textbf{Task}}{\textbf{Item}}}& \textbf{Few-Shot}  & \makecell[c]{\textbf{Sequential} \\ \textbf{task}} &  \makecell[c]{\textbf{Forgetting} \\ \textbf{alleviation}} & \makecell[c]{\textbf{Class} \\ \textbf{increment}} & \makecell[c]{\textbf{Domain} \\ \textbf{increment}} 
 & \makecell[c]{\textbf{Overall evaluation} \\ \textbf{(seen domain)}}   & \makecell[c]{\textbf{Prediction on} \\ \textbf{(unseen domain)}} \\
\bottomrule[1.3pt]
Finetuning & & \XSolidBrush & \XSolidBrush & \XSolidBrush & \XSolidBrush & \Checkmark & \XSolidBrush \\
\cline{1-8} 
Domain Adaptation &\XSolidBrush & \XSolidBrush & \XSolidBrush& \XSolidBrush & \XSolidBrush & \Checkmark & \XSolidBrush \\
Domain Generalization & \XSolidBrush & \XSolidBrush&  \XSolidBrush & \XSolidBrush & \XSolidBrush & \XSolidBrush & \Checkmark  \\
Few-Shot Learning & \Checkmark & \XSolidBrush & \XSolidBrush & \XSolidBrush & \XSolidBrush  & \Checkmark &  \XSolidBrush \\
\cline{1-8} 
Class-Incremental Learning & \XSolidBrush & \Checkmark & \Checkmark & \Checkmark &  \XSolidBrush & \Checkmark &  \XSolidBrush \\
Domain-Incremental Learning~\cite{wang2022s} & \XSolidBrush & \Checkmark & \Checkmark & \XSolidBrush &  \Checkmark & \Checkmark &\XSolidBrush \\
Cross-Domain Continual Learning~\cite{simon2022generalizing} & \XSolidBrush & \Checkmark & \Checkmark &  \Checkmark & \XSolidBrush & \XSolidBrush & \Checkmark \\
General Incremental Learning~\cite{xie2022general} &\XSolidBrush & \Checkmark & \Checkmark & \Checkmark  & \Checkmark &  \Checkmark & \XSolidBrush\\
Continual Domain Adaptation & \XSolidBrush & \Checkmark &\Checkmark & \XSolidBrush &\Checkmark & \Checkmark & \XSolidBrush \\

\cline{1-8} 
Few-Shot Class-Incremental Learning~\cite{tao2020few} & \Checkmark & \Checkmark & \Checkmark&  \Checkmark &\XSolidBrush & \Checkmark & \XSolidBrush  \\
GFSCL (Ours) &\Checkmark & \Checkmark & \Checkmark &\Checkmark & \Checkmark & \Checkmark & \Checkmark \\
\bottomrule[1.3pt]
\end{tabular}
}
\label{table:related}
\end{table*}

For domain-related issues, the model is required to preserve the knowledge and capture the domain-invariant representation for generalization. In this way, prototype-calibrated contrastive learning is put forward. Since our framework is rehearsal-free, we only store the prototype for a specific category (\ie, the mean of all previous training samples of this class in the feature space), which can protect data privacy and is memory-efficient. In detail, we alleviate the intra-class variation by decreasing the distance between the class prototype obtained in previous domains and features of new domains with contrastive loss. Our contributions include:
\begin{itemize}
\setlength\itemsep{-0.1em}
    \item Under the limited data setting, we are the first to explore the GFSCL configuration by considering class- and domain-incremental tasks together with the assessment of domain generalization capability simultaneously, which is practical and challenging. Benchmark datasets and protocols are provided with exhaustive baselines in this paper. 
    \vspace{0.6em}
    \item To deal with GFSCL, we design the Contrastive Mixtures of Adapters (CMoA) based on the ViT-based rehearsal-free framework, which can efficiently fulfill the stability-plasticity dilemma while freezing the majority of ViT parameters. CMoA includes Mixtures of Adapters (MoA) for the class-incremental purpose and prototype-calibrated contrastive learning for domain-related issues. 
    \vspace{0.6em}
    \item Finally, extensive experiments are conducted on the two arranged datasets to validate the efficacy of CMoA towards comprehensive evaluation indicators. The results demonstrate that CMoA exceeds other rehearsal-free continual learning methods, and achieves comparative performance with rehearsal-based ones.  
    
\end{itemize}

\section{Related Work}
\label{sec:relatedwork}
In this section, we provide a brief summary of the related research topics in Table~\ref{table:related}. We also introduce the recent progress in few-shot learning. Then, some works on continual learning are investigated. Finally, related domain generalization methods are presented.

\subsection{Few-Shot Learning (FSL)}
Few-Shot Learning (FSL)~\cite{snell2017prototypical, finn2017model, wang2020instance, munkhdalai2017meta, sun2019meta, perez2021true, lee2019meta, vinyals2016matching} is an emerging research topic that aims to first learn with base classes and then adapt to disjoint new classes with limited training samples. There are mainly two directions to tackle this problem. One is optimization-based methods, which firstly train a network with base class data, then finetune the network with new support data. MAML \cite{finn2017model} aimed at obtaining optimal initialization parameters of the model for a novel task through meta-training. MetaNet~\cite{munkhdalai2017meta} proposed by introducing fast and slow
weights. MetaOptNet~\cite{lee2019meta} learned feature embeddings that generalize well under a linear classification rule for novel categories. Metric-based methods learn a semantic embedding space and classify query samples based on
their similarity. Metrics such as cosine similarity~\cite{vinyals2016matching}, Euclidean distance~\cite{snell2017prototypical}, Mahalanobis distance~\cite{bateni2020improved}, and Earth
Mover’s Distance (EMD)~\cite{zhang2020deepemd} have been effectively applied to FSL. CrossTransformers~\cite{doersch2020crosstransformers} explored coarse spatial correspondence between the query and the labeled images and then used the spatially-corresponding features for classification.

\begin{figure*}[t]
\includegraphics[width=0.9\linewidth]{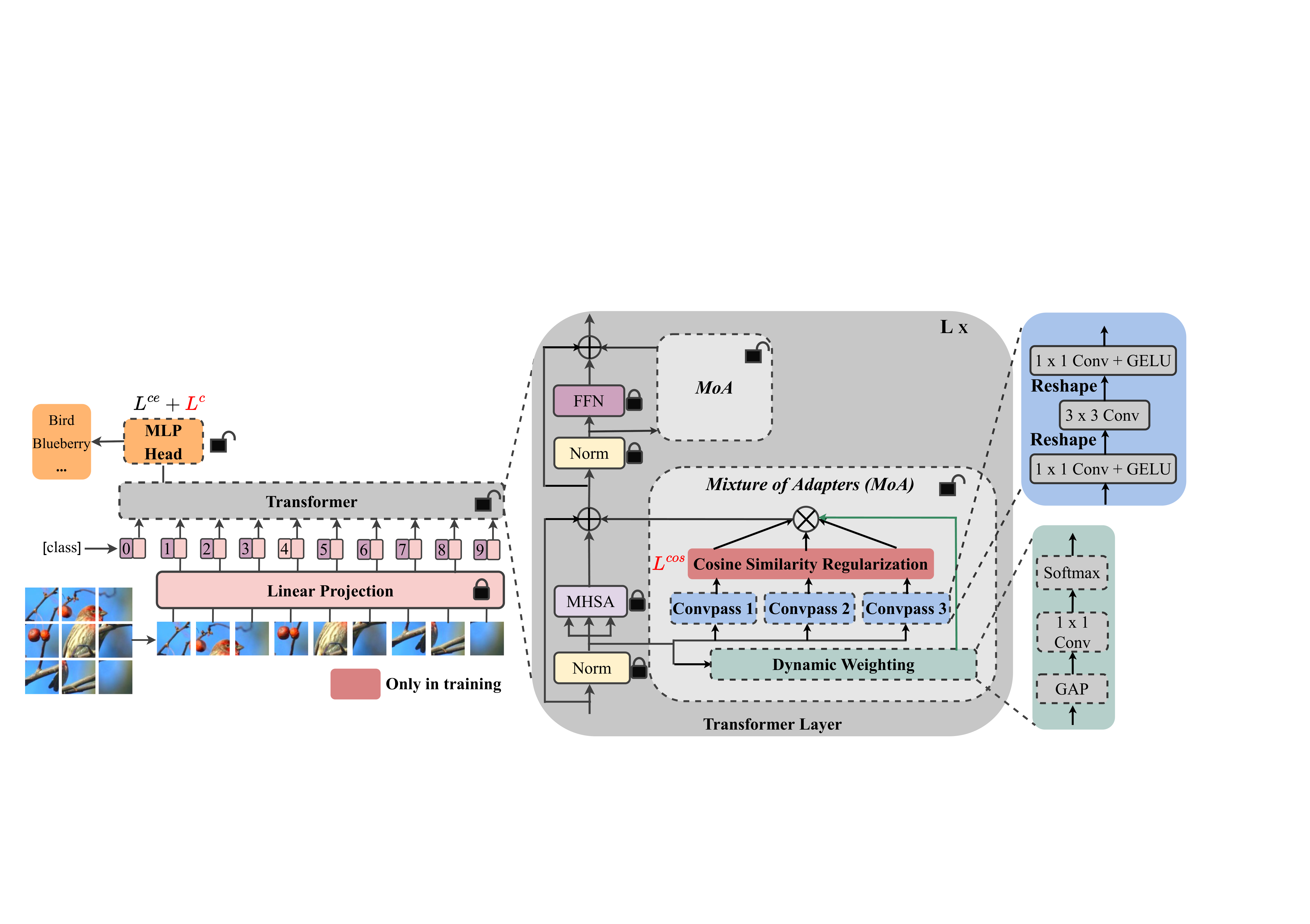}
\centering
\caption{The framework of the CMoA. Before training, we load the pre-trained model of ViT and combine two MoA modules into each transformer block. We put image patches and an extra learnable classification embedding into the transformer. From the first session, during the training stage, we freeze the parameters of linear projection layer and vanilla transformer blocks while leaving MoA module trainable. Besides, the MLP head is expanded to adjust to the increase of novel classes. During testing, the cosine similarity regularization section is not required since the regularization aims to guide each adapter to learn different knowledge and focus on different classes. `MHSA', `FFN', and `GAP' are short for the multi-head self-attention, feed-forward network, and global average pooling.}
\label{method} 
\end{figure*}
 
\subsection{Continual Learning}
Recently, there are three main series of
continual learning algorithms. Regularization-based~\cite{kirkpatrick2017overcoming, aljundi2018memory, li2017learning}
methods aim to alleviate catastrophic forgetting by limiting the learning rate on important parameters for previous tasks. MAS~\cite{aljundi2018memory} computed the importance of the parameters of a neural network in an unsupervised and online manner, then changes to important parameters can then be penalized. Rehearsal-based methods~\cite{rebuffi2017icarl, riemer2018learning,hu2019overcoming} reserved a data buffer for saving previous samples from older tasks to train with data
from the current task. iCaRL~\cite{rebuffi2017icarl} learned strong classifiers and a data representation simultaneously. Distillation-based methods~\cite{hou2019learning, wu2019large, tao2020topology} apply the technique of knowledge distillation to mitigate catastrophic forgetting. Hou~\etal incorporated three components to mitigate the adverse effects of the data imbalance between previous and new data.

\subsection{Few-Shot Class-Incremental Learning}
Following CIL, FSCIL concentrates on the challenging problem of learning the streaming novel tasks with few samples provided. The main issue of FSCIL is the trade-off between learning new knowledge and preventing forgetting past knowledge. Tao \etal~\cite{tao2020few} benchmark FSCIL and propose TOPIC to model the topology of the feature space using neural gas. Zhang \etal~\cite{zhang2021few} employ a simple but effective decoupled learning strategy of representations, and Continually Evolved Classifier (CEC) is proposed by using a graph model to propagate context information between classifiers for adaptation. Zhu \etal ~\cite{zhu2021self} offer a novel incremental prototype learning scheme to solve the FSCIL task. Knowledge distillation was also applied to prevent catastrophic forgetting by using the previous model as the teacher model~\cite{cheraghian2021semantic,dong2021few}. Zhou \etal ~\cite{zhu2021self} improve the FSCIL performance by preserving areas on embedding space for upcoming classes. Shi~\etal~\cite{shi2021overcoming} constrain the parameter updating at the base session and fine-tune the model parameters within the flat local minima of the base training objective function.

\subsection{Domain Generalization}
Domain generalization aims to learn a model from
multiple source domains such that the model can generalize on the unseen target domain. Existing works mainly follow three directions: (1) enriching data diversity~\cite{xu2021fourier, zhou2021domain}. Xu~\etal~\cite{xu2021fourier} propose a novel Fourier-based data augmentation strategy called amplitude mix and a dual-formed consistency loss called co-teacher regularization is further introduced. (2) Obtaining domain-invariant representations or domain-specific features~\cite{piratla2020efficient, carlucci2019domain}. CSD~\cite{piratla2020efficient} jointly learns a common component and a domain-specific component. (3) Exploiting
general learning strategies~\cite{li2019episodic, huang2020self}. Huang~\etal~\cite{huang2020self} introduce a simple training heuristic, Representation Self-Challenging (RSC) to improve the generalization ability.

\section{Methodology}
\label{sec:method}
In this section, we first introduce the formulation of generalized few-shot continual learning in Section~\ref{sec:setup}. To enhance the class-incremental learning ability, we then propose a mixture of adapters in  Section~\ref{sec:moa}. Finally, to improve domain-related issues, we present the prototype-calibrated contrastive learning in  Section~\ref{sec:pcl}.

\subsection{Problem Setup}
\label{sec:setup}
Firstly, we define a sequence of tasks as 
$\mathcal{T}=\{t_1, t_2, ..., t_n\}$, and the dataset sequence is $\mathcal{D}=\{\mathcal{D}_1, \mathcal{D}_2, ...,\mathcal{D}_n\}$. For a specific dataset $\mathcal{D}_i$ in session $i$, category sets and the domain are denoted as $C_i$, and $d_i$. For GFSCL, in two neighboring sessions, $\mathcal{C}_{i} \cap \mathcal{C}_{i+1} = \emptyset$ or $d_i \neq d_{i+1}$, which means that the model conducts class-incremental or domain-incremental task in each incremental learning session. Here $\mathcal{D}_1$ is the large-scale base dataset used in the first base session, and the followings are all novel few-shot datasets. Specifically, we term $\mathcal{D}_{1}=\left\{\left(\bm{x}_j, y_j\right)\right\}_{j=1}^{\lvert  \mathcal{D}_1\rvert}$ where $ y_j \in \mathcal{C}_1$, and $\mathcal{C}_1$ represents the base category set. In the $i$-th session where $i>1$, the novel/new dataset is defined as $\mathcal{D}_{i}=\left\{\left(\bm{x}_j, y_j\right)\right\}^{N \times K}_{j=1}$ consists of $N$ classes $\mathcal{C}_i$ with $K$ labeled examples per class, \ie, a $N$-way $K$-shot problem. It is worth noting that there is no overlap between samples of different sessions, \ie, $\mathcal{D}_{i} \cap \mathcal{D}_{i'} = \emptyset$. Moreover, there is also no overlap between the categories of different class-incremental sessions, \ie, $\mathcal{C}_{i} \cap \mathcal{C}_{i'} = \emptyset$ and $\mathcal{D}_{i} \cap \mathcal{D}_{i'} = \emptyset$, where $i\neq i'$ and session $i$ and session $i'$ are all class-incremental sessions.

In this paper, we arrange the GFSCL configuration and validate the efficacy of the proposed CMoA illustrated in Fig.~\ref{method} on the object classification task. The objective of GFSCL is to cumulatively learn novel classes or old classes in a new domain, and then the final overall classification accuracy will be evaluated on all seen classes and domains. Meanwhile, the generalization ability on new domains is also evaluated.

\subsection{Mixture of Adapters}
\label{sec:moa}
One of the tasks in GFSCL is class-incremental learning, and the target is to continually learn novel categories without forgetting previously seen categories. Due to the limited data regime, we incorporate adapters into each transformer layer for the fast adaption purpose, which is suitable for solving few-shot learning tasks~\cite{bansal2022meta, li2022cross, zhang2022tip}. However, the performance of the previous task is difficult to maintain when the model adapts fast to novel tasks, because conducting the fast adaption may require a larger learning rate and stronger gradients from
new classes’ classification loss~\cite{tao2020few}. 
To solve the above issue, we incorporate the MoA module into each transformer block illustrated in Fig.~\ref{method} to replace the single adapter. These incorporated MoA modules are employed as the adaptation module in each transformer block for the fast adjustment to newly encountered tasks, and we hope that each adapter can focus on different classes for stability purpose. For the mixture aim, there are two components in MoA: cosine similarity regularization and dynamic weighting strategy. In MoA module, we choose Convpass~\cite{jie2022convolutional} as each adapter in MoA module because of the hard-coded inductive bias of convolutional layers in Convpass~\cite{jie2022convolutional} that is more suitable for visual tasks. 

%For original ViT with Convpass, Convpass is an inserted convolutional bottleneck block parallel to the MHSA or MLP block as an adaptation module in each transformer block.
%Furthermore, the MoA module can also tackle the issue of learning with limited data since only the parameters of this ensemble module and MLP head are finetuned for fast adaptation when the novel task arrives. 

\vspace{0.2em}
\noindent\textbf{Cosine similarity regularization.} When conducting experiments, we find that different adapters tend to learn repetitive information; thus it leads to limited performance enhancement. Hence, we involve a cosine similarity regularization into outputs of $P$ adapters to lead them to learn different knowledge. Specifically, in $q$-th MoA module of transformer layer $l$, the input of the MoA module is the output of layer normalization and is denoted as $\boldsymbol{A}_{l,q}$. Moreover, we define the output of adapter $j$ as $\boldsymbol{H}_j$ and the output of the MoA module as $\boldsymbol{B}_{l,q}$. In order to learn diverse features and obtain complementary knowledge, we add cosine similarity regularization into the outputs of adapters. In detail, For each pair of outputs of adapters $j$ and $j'$ where $j\neq j'$, we minimize the cosine similarity between $\boldsymbol{H}_j$ and $\boldsymbol{H}_{j'}$. Assuming that the feature dimension is $D$ and there are $M$ tokens in the input image, the final cosine distance is computed along the feature dimension and by averaging over the number of tokens. Commonly, the cosine similarity loss is defined as 
\begin{equation}\small
\mathcal{L}^{cos}_{l,q} =  \sum_{1 \le j, j' \le P, j\ne j'}^{} \frac{1}{M} \sum_{m=1}^{M}\left ( \frac{\boldsymbol{H}_{j,m}\cdot \boldsymbol{H}_{j',m}}{\left \| \boldsymbol{H}_{j,m} \right \|  \left \| \boldsymbol{H}_{j',m} \right \|}  \right )^2.
\label{eq:l_cos}
\end{equation}

However, the regularization implemented by Eq.~\ref{eq:l_cos} is too strict in the limited data setting. For this reason, we propose to apply the loose regularization by defining the cosine similarity loss as
\begin{equation}\footnotesize
\mathcal{L}^{cos}_{l,q} =  \sum_{1 \le j,j'\le P,j\ne j'}^{} \frac{1}{M} \sum_{m=1}^{M} \bm{{\rm max}}\left ( \frac{\boldsymbol{H}_{j,m}\cdot \boldsymbol{H}_{j',m}}{\left \| \boldsymbol{H}_{j,m} \right \| \left \|\boldsymbol{H}_{j',m}\right \|} - \gamma, \ 0\right ),
\label{eq:l_cos_loose}
\end{equation}
where $\gamma$ is a hyper-parameter that controls the strength of the cosine similarity regularization. There are totally $L$ transformer layers, and the total cosine similarity loss is defined as 
\begin{equation}\small
\mathcal{L}^{cos} =  \sum_{l=1}^{L} \sum_{q=1}^{2}\mathcal{L}^{cos}_{l,q},
\label{eq:total_cos}
\end{equation}
where 2 is the number of MoA modules in each transformer layer, which is the same as the number of the adapters~\cite{houlsby2019parameter, jie2022convolutional}.

\vspace{0.2em}
\noindent\textbf{Dynamic weighting.} To guide each adapter to concentrate on different categories, we propose a dynamic weighting strategy for the mixture procedure of multiple adapters. The dynamic weighting strategy consists of one group average pooling layer and one $1 \times 1$ convolutional layer followed by the softmax function. Here the dynamic weighting strategy is to assign weights dynamically to adapters, and this strategy is defined as $\phi \left ( \cdot \right)$. The dynamic weight set $\mathcal{W}$ generated by
\begin{equation}
\mathcal{W}_{l,q} = \phi \left ( \boldsymbol{A}_{l,q}  \right ) 
\label{eq:adaptive}
\end{equation}
where $\mathcal{W}_{l,q} = \{w_1, w_2,... , w_P\}$. The final output of MoA $\boldsymbol{B}$ is obtained by the following aggregation function:
\begin{equation}
\boldsymbol{B}_{l,q} = \sum_{i=1}^{P} w_i \cdot  \boldsymbol{H}_i.
\label{eq:aggregation}
\end{equation}
After the aggregation process, $\boldsymbol{B}$ is forwarded to the next layer.

In the class-incremental session including the first session, the total loss is defined as 
\begin{equation}
\mathcal{L}^t = \mathcal{L}^{ce} + \zeta\mathcal{L}^{cos},
\label{eq:total-loss-class}
\end{equation}
where $\mathcal{L}^{ce}$ is the cross-entropy loss and $\zeta$ is the hyper-parameter representing the weight of $\mathcal{L}^{cos}$. 

\begin{figure}[t]
\includegraphics[width=1.0\linewidth]{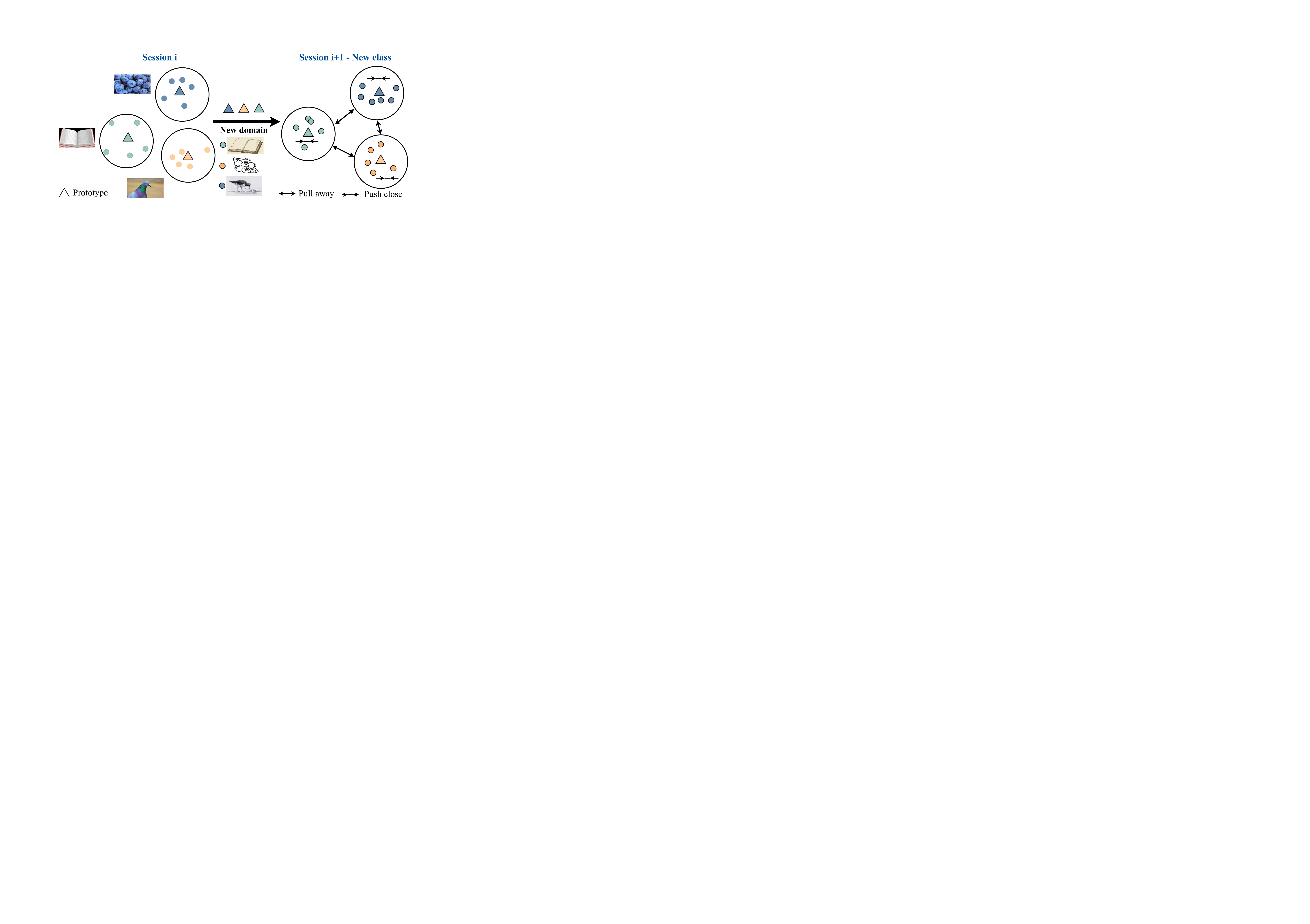}
\centering
\caption{The operation of prototype-calibrated contrastive learning.}
\label{figure:contrastive} 
\end{figure}

\subsection{Prototype-Calibrated Contrastive Learning}
\label{sec:pcl}
For GFSCL, the domain-related issues involve domain-incremental learning and generalization assessment in unseen domains. Hence, the model is required to maintain the performance of previously seen domains and improve the generalization ability in unseen domains. To achieve this, we propose prototype-calibrated contrastive learning illustrated in Fig.~\ref{figure:contrastive} to assist the model in learning domain-invariant representation. 

Our proposed CMoA is a rehearsal-free framework in which the samples of previous tasks do not need to to stored in extra memory for further finetuning in the following incremental sessions, which is beneficial for protecting privacy. Moreover, all the features of training samples are not required to preserve in our proposed method, which is memory-efficient. Alternatively, we store the prototypes of previously seen classes, \ie, one prototype for a certain class. The prototype of a specific class can be regarded as the centroid of this class in the feature space and is defined by averaging the obtained training features. For session $i$, if it is a class-incremental session, the prototypes of the newly encountered classes are termed $\mathbf{\bar{P}}_i = \left \{\mathbf{\bar{p}}_{i,j} \right \}_{j=1}^{N}$ and $\mathbf{\bar{p}}_{i,j}$ is computed by
\begin{equation}
\mathbf{\bar{p}}_{i,j} = \frac{1}{K} \sum_{k=1}^{K} \Theta\left ( \bm{x}_k \right ), 
\label{eq:prototype}
\end{equation}
where $\Theta\left ( \cdot  \right ) $ is the backbone. If it is a domain-incremental session, the prototypes are updated by averaging previous prototypes and obtained training features in the current domain:
$\mathbf{\bar{p}}_{i,j}$ is computed by
\begin{equation}\small
\mathbf{\bar{p}}_{i,j} = \frac{1}{(E+1)K} \left (\sum_{k=1}^{K} \Theta\left ( \bm{x}_k \right ) + E \times K \cdot \mathbf{\bar{p}}_{i-1,j}\right ), 
\label{eq:prototype}
\end{equation}
where $E$ is the number of previous domains that these classes encountered.

\begin{table*}[t]
\caption{Protocol of \textit{mini}-DomainNet. The unseen domain is \textit{Painting}.}
%Our proposed UaD-CE achieves the state-of-the-art results with respect to the three indicators.
\centering
\setlength{\tabcolsep}{11pt}
\resizebox{0.8\textwidth}{!}{
\tabcolsep 0.06in
\begin{tabular}{c|ccccccccc}
\bottomrule[1.3pt]
{\diagbox{\textbf{Item}}{\makecell[c]{\textbf{Session} \\ \textbf{ID}}}}& \textbf{1}  & \textbf{2} &\textbf{3}
 &\textbf{4} & \textbf{5}& \textbf{6}&\textbf{7} & \textbf{8}& \textbf{9}\\
\bottomrule[1.3pt]
Task & CIL, DIL & CIL & DIL & CIL & DIL & CIL & DIL &  CIL & DIL \\
\cline{1-10} 
 Domain & Real & Real & Clipart & Clipart & Sketch & Sketch & Infograph & Infograph & Quickdraw  \\
\cline{1-10} 
Class ID & 0-59 & 60-64 & 60-64 & 65-69 & 65-69 & 70-74 & 70-74 & 75-79 & 75-79\\
\cline{1-10} 
Dataset & Large & \makecell[c]{5-Way \\ 5-Shot} & \makecell[c]{5-Way \\ 5-Shot} & \makecell[c]{5-Way \\ 5-Shot} & \makecell[c]{5-Way \\ 5-Shot} & \makecell[c]{5-Way \\ 5-Shot} & \makecell[c]{5-Way \\ 5-Shot} & \makecell[c]{5-Way \\ 5-Shot} & \makecell[c]{5-Way \\ 5-Shot} \\
\bottomrule[1.3pt]
\end{tabular}
}
\label{table:domainet-protocol}
\end{table*}

To be specific, the contrastive loss is implemented by pair-wise cross-entropy loss. Firstly, we sample $S$ positive pairs belonging to different classes from $\mathbf{\bar{P}}_i \cup \mathcal{D}_{i+1}$, and each positive pair $(\bm{x}_m,\bm{x}_n)$ contains two samples from the same category. Generally, $S$ is  equal to the training batch size for every epoch. The total number of samples is $T=2S$. Here the loss function for a specific positive pair is defined as 
\begin{equation}\small
\mathcal{L}^c_{m,n} = -log\frac{exp(-d(\Theta\left ( \bm{x}_m \right ), \Theta\left ( \bm{x}_n \right ))/\tau )}{ {\textstyle \sum_{t=1,t\ne m}^{T}} exp(-d(\Theta\left ( \bm{x}_m \right ), \Theta\left ( \bm{x}_t \right ))/\tau)},  
\label{eq:metric_loss}
\end{equation}
where $\tau$ is the temperature hyper-parameter and $d(\cdot)$ is the distance function. Commonly, the aim of contrastive learning is to push the representations of the same class to be closer and others to be farther. In the domain-incremental session $i+1$, the prototypes of the previous session stand for the old knowledge of the previous-seen domain. In our prototype-calibrated contrastive learning, we incorporate an extra contrastive part by forcing the training features of the new domain to be closer to the corresponding prototype and to be farther from the prototypes of other categories for memorizing the old domains. For a particular class, by narrowing the distance among seen domains in the feature space, it can instruct the model to learn domain-invariant representation, which can further improve the performance in unseen domains. 

In the domain-incremental session except for the first session, we define the total loss as 
\begin{equation}
\mathcal{L}^t = \mathcal{L}^{ce} + \mathcal{L}^{c},
\label{eq:total-loss-domain}
\end{equation}
where $\mathcal{L}^{c}$ is the contrastive loss of all positive pairs.

\section{Experiments}
\label{sec:experiment}

\subsection{Datasets and Evaluation Metrics}
In this paper, we benchmark two protocols for GFSCL based on DomainNet~\cite{peng2019moment} and ImageNet-C~\cite{hendrycks2018benchmarking}. We name the two newly arranged datasets as \textit{mini}-DomainNet and \textit{mini}-ImageNet-C.

\begin{table*}[t]
\caption{Protocol of \textit{mini}-ImageNet-C. \textit{GN}, \textit{DB}, \textit{B} mean gaussian noise, defocus blur, and brightness, respectively. The unseen domains are Impulse Noise, Zoom Blur, JPEG.}
%Our proposed UaD-CE achieves the state-of-the-art results with respect to the three indicators.
\centering
\setlength{\tabcolsep}{11pt}
\resizebox{\textwidth}{!}{
\tabcolsep 0.06in
\begin{tabular}{c|ccccccccccccc}
\bottomrule[1.3pt]
{\diagbox{\textbf{Item}}{\makecell[c]{\textbf{Session} \\ \textbf{ID}}}}& \textbf{1} & \textbf{2} &\textbf{3}
 &\textbf{4} & \textbf{5}& \textbf{6}&\textbf{7} & \textbf{8}& \textbf{9} & \textbf{10} & \textbf{11} & \textbf{12} & \textbf{13} \\
\bottomrule[1.3pt]
Task & CIL, DIL & CIL & DIL & DIL & CIL & DIL & DIL & CIL & DIL & DIL & CIL & DIL & DIL \\
\cline{1-14} 
Domains & \makecell[c]{GN, DB \\ Fog, B} & GN & Pixelate  & \makecell[c]{Frosted \\ Glass Blur} & \makecell[c]{Frosted \\ Glass Blur} & Elastic & \makecell[c]{Shot \\ Noise}  & \makecell[c]{Shot \\ Noise} & Snow & Contrast & Contrast & \makecell[c]{Motion \\ Blur} & Fog \\
\cline{1-14} 
Class ID & 0-59 & 60-69 & 60-69 & 60-69 & 70-79 & 70-79 & 70-79 & 80-89 & 80-89 & 80-89 & 90-99& 90-99& 90-99 \\
\cline{1-14} 
Dataset & Large-scale & \makecell[c]{10-Way \\ 5-Shot} &\makecell[c]{10-Way \\ 5-Shot} & \makecell[c]{10-Way \\ 5-Shot} & \makecell[c]{10-Way \\ 5-Shot} &\makecell[c]{10-Way \\ 5-Shot} & \makecell[c]{10-Way \\ 5-Shot}& \makecell[c]{10-Way \\ 5-Shot} &
\makecell[c]{10-Way \\ 5-Shot} &
\makecell[c]{10-Way \\ 5-Shot} & \makecell[c]{10-Way \\ 5-Shot} & \makecell[c]{10-Way \\ 5-Shot} & \makecell[c]{10-Way \\ 5-Shot} \\
\bottomrule[1.3pt]
\end{tabular}
}
\label{table:imagenet-protocol}
\end{table*}

\begin{figure*}[t]
\vspace{-1.2em}
\includegraphics[width=0.8\linewidth]{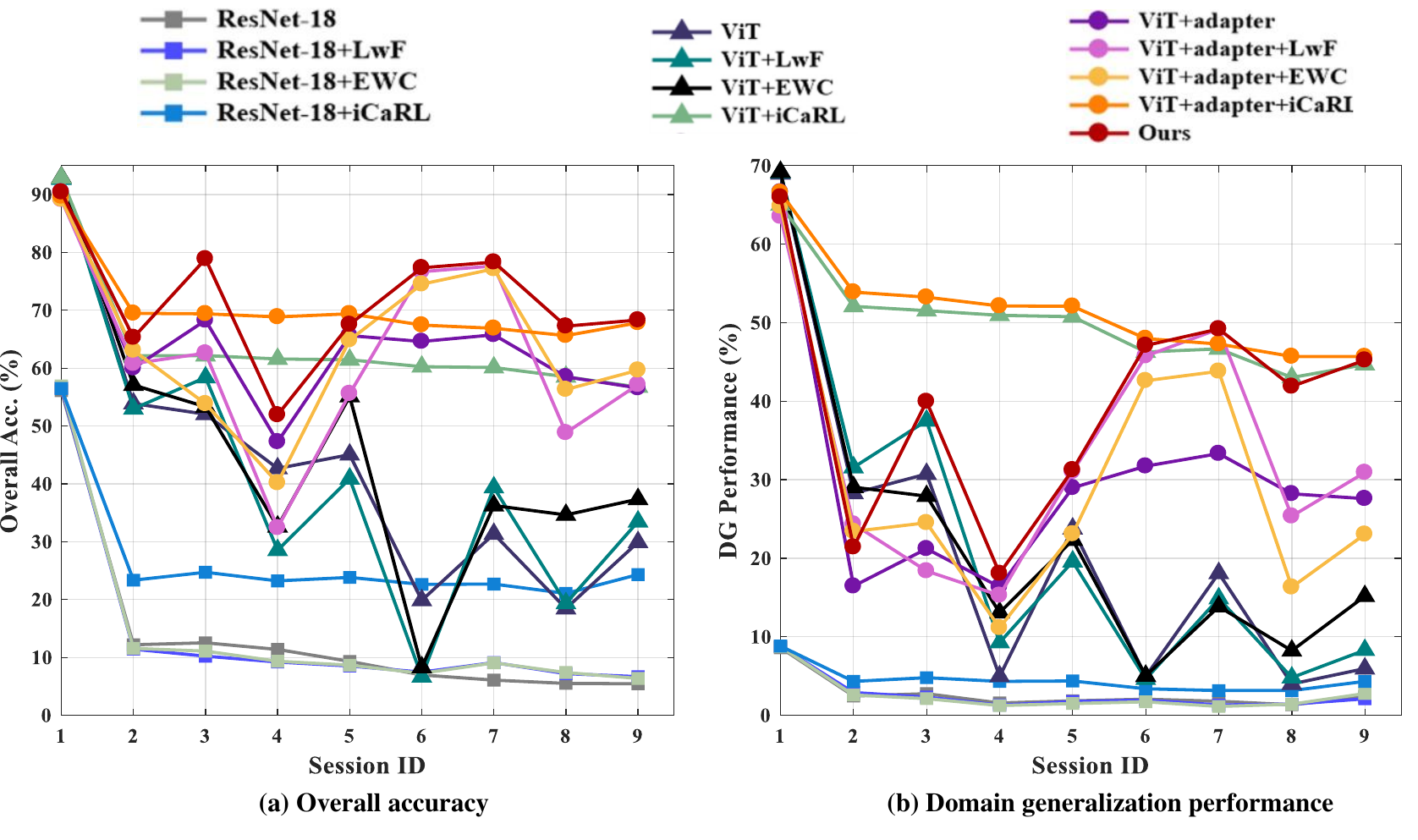}
\centering
\caption{Baselines and comparative study with \textit{mini}-DomainNet. iCaRL~\cite{rebuffi2017icarl} is the rehearsal-based continual learning method.}
\label{domainnet} 
\end{figure*}

\begin{table*}[t]
\caption{Comparative study with \textit{mini}-ImageNet-C. $iCaRL^{*}$~\cite{rebuffi2017icarl} is the rehearsal-based continual learning method. \textit{Acc.} means the overall classification test on all seen classes and domains. \textit{DG.} stands for the generalization ability in unseen domains. \textit{Average Acc.} means the average performance on seen classes and seen domains of all sessions. \textit{Average DEG} represents the average domain generalization enhancement.}
%Our proposed UaD-CE achieves the state-of-the-art results with respect to the three indicators.
\centering
\setlength{\tabcolsep}{11pt}
\resizebox{15.5cm}{!}{
\tabcolsep 0.06in
\begin{tabular}{ccccccccccccccccc}
\bottomrule[1.3pt]
 \multirow{2}{*}{\textbf{Method}}&  \multirow{2}{*}{\textbf{Eval.}}& \multicolumn{13}{c}{\textbf{Session ID}} & \multirow{2}{*}{\makecell[c]{\textbf{Average} \\ \textbf{Acc.}}} & \multirow{2}{*}{\makecell[c]{\textbf{Average} \\ \textbf{DGE}}}\\ 
\cline{3-15} 
 & &  \textbf{1} & \textbf{2} &\textbf{3}
 &\textbf{4} & \textbf{5}& \textbf{6}&\textbf{7} & \textbf{8}& \textbf{9}& \textbf{10} &\textbf{11} & \textbf{12} & \textbf{13}\\
\bottomrule[1.3pt]
& \multicolumn{16}{c}{\cellcolor[HTML]{EFEFEF} \textbf{ResNet-18}} \\
\cline{2-17}  
\multirow{2}{*}{Finetune} & Acc. & 56.34 & 17.76 & 16.97 & 17.04 & 14.02 & 15.24 & 15.63& 13.30 & 13.42 & 14.79 & 13.70 & 13.70 & 13.84 & 18.13 & -\\

& DG & 38.62 & 14.58 & 13.94 & 13.97 & 13.90 & 11.57 &12.04 & 8.36 & 9.51 & 9.93 &9.04 & 9.21 & 4.66 &- & -1.32 \\
\cline{2-17} 
 \multirow{2}{*}{EWC~\cite{kirkpatrick2017overcoming}} & Acc. & 57.23 & 18.24 & 18.18 & 18.02 & 16.05 & 17.45 & 16.67 & 14.27 & 16.24 & 14.95 & 14.04 & 15.21 & 16.01 & 19.43 &-\\
& DG & 37.67 & 17.55 & 17.31 & 17.02 & 15.03 & 17.39 & 15.29 & 13.59 & 14.66 & 13.71 & 13.18 & 14.16 & 14.03 & - & 0.18 \\
 \cline{2-17} 
\multirow{2}{*}{LwF~\cite{li2017learning}} & Acc. & 56.34 & 18.93 & 15.85 & 17.06 & 14.80 & 16.03 & 14.99 & 13.33 & 15.13 & 13.93 & 13.96 & 14.77 & 14.70 & 18.45 & -\\
& DG & 37.44 & 15.44 & 15.97 & 14.73 & 14.00 & 15.53 & 15.29 & 12.94 & 13.79 & 12.90 & 12.88 & 12.90 & 13.61 & - & 0.33\\
\cline{2-17}
\multirow{2}{*}{$iCaRL^{*}~\cite{rebuffi2017icarl}$} & Acc. & 56.24 & 33.42 & 33.63 & 32.47 & 31.53 & 29.61 &29.83 &29.15 & 31.07 & 30.37 &29.14 &29.71 & 30.22 & 32.80&-\\ 
& DG & 37.12 & 20.51 & 19.66 & 19.56 &17.00 & 15.70 & 16.74 & 15.79 &16.03 &15.07 &14.22 &14.58 &14.27 & -& -0.47\\
\bottomrule[1.3pt]
& \multicolumn{16}{c}{\cellcolor[HTML]{EFEFEF}\textbf{ViT}} \\
\cline{2-17}
 \multirow{2}{*}{Finetune}  & Acc. & 89.01 &59.28 &60.85 &56.43 & 28.39& 32.27& 34.76& 29.92& 35.94& 37.28& 47.97& 50.32& 49.19& 47.03 & - \\
& DG & 88.29 & 44.04 & 45.36 & 41.89 & 26.61 &29.03& 30.71& 23.12& 28.73& 29.53& 39.92& 41.51& 41.98& - & 2.60\\
\cline{2-17}
\multirow{2}{*}{EWC~\cite{kirkpatrick2017overcoming}} & Acc.& 89.26 &54.69& 57.16 &51.66& 38.83& 38.98& 41.82& 14.89& 23.26& 29.56 &38.66 &43.54& 51.65&  44.15 & -  \\
 & DG & 88.75 &46.95& 51.61& 47.04& 38.48& 36.10& 36.25& 23.41& 19.09& 13.00& 30.79& 36.95& 33.53 & - &  -2.32\\
 \cline{2-17}
 \multirow{2}{*}{LwF~\cite{li2017learning}} & Acc. & 89.08 & 49.73 & 52.45 & 50.89 & 41.55 & 44.56 & 45.53 & 34.03 & 35.57 & 40.77& 40.88 &46.22 &46.28 & 47.50 & -\\
& DG & 89.05 & 47.89& 47.56 & 45.14 & 39.12 & 42.01& 42.45& 33.59& 30.10& 28.50 & 39.92& 40.34& 33.40& - & -2.76 \\
\cline{2-17}
 \multirow{2}{*}{$iCaRL^{*}~\cite{rebuffi2017icarl}$}& Acc. &  89.91 &67.30& 66.61&65.86&63.85&62.55&61.25&59.39&58.39 &57.76&57.15&56.71&56.82 & 63.35 & - \\ 
 & DG & 88.65&53.54&53.34&53.01&48.41&48.01&46.85&42.94&42.21&41.81&39.41&41.56&45.44& -& 0.70
\\
 
\bottomrule[1.3pt]
& \multicolumn{16}{c}{\cellcolor[HTML]{EFEFEF}\textbf{ViT + Adapter}} \\
\cline{2-17}
\multirow{2}{*}{Finetune} & Acc. & 91.25 & 41.80 & 50.77 & 54.64&  40.19 & 50.10 & 55.75 & 55.34 &54.89 &55.39 &51.35 &56.20 &57.77 & 55.03 & - \\
& DG & 82.25 & 33.56 &41.23 &39.54 &26.37 &33.67 &38.46 &36.78 &36.86 &37.18 &36.10 &39.59 &39.95 & - & 5.58 \\
\cline{2-17}
\multirow{2}{*}{EWC~\cite{kirkpatrick2017overcoming}} & Acc. & 91.60 & 55.12 & 57.89 &50.63 & 46.91 & 54.96 & 51.42 &  52.84 & 52.14 & 51.32& 54.20& 59.85& 60.74 & 56.89 &-\\
 
 & DG &82.74 & 43.65 & 46.39 & 38.97 & 30.88 & 37.17& 42.72 &44.92 & 44.78 & 43.91 & 39.21 & 42.88 & 42.69 & - & 2.41\\
 \cline{2-17}
\multirow{2}{*}{LwF~\cite{li2017learning}} & Acc.  &  91.82  & 45.24 & 57.01 & 55.56 & 28.05 & 40.86 & 49.69 & 36.76 & 47.60 & 54.43 & 54.49 & 59.20 & 60.49 & 52.40 & - \\
& DG & 82.62 & 30.59& 37.04& 35.89& 28.75 & 28.54& 31.72& 30.95& 30.35& 33.40& 36.34 & 40.84 & 40.82 & -& 3.80  \\
\cline{2-17}
 \multirow{2}{*}{$iCaRL^{*}~\cite{rebuffi2017icarl}$} & Acc. & 91.40 &67.96 &69.15 & 68.41 & 67.18 & 66.77 & 66.29 & 65.16 & 64.92 & 64.81 & 64.24 & 64.16&  \textbf{64.19} & \textbf{66.57} & -\\ 
& DG & 82.50 & 48.67 & 56.33& 58.04 & 48.61 & 51.60 &54.26 &51.82 &52.01 &52.06 &51.14 &51.96 & \textbf{51.26} & -& 3.85 \\
\cline{2-17}
\multirow{2}{*}{\makecell[c]{\textbf{CMoA} \\ \textbf{(Ours)}}} & Acc. & 91.80 & 53.52& 62.06&49.98 & 51.26 &57.27 &61.03 &68.07 &66.47 &66.10 &61.08& 63.41& 64.10 & 62.78 & -\\
&DG & 82.67 & 31.48 &42.44 &35.44 &34.91 &39.86 &43.47 &40.92 &49.21 &50.28 &43.62 & 46.23 & 46.10 &- & \textbf{6.09}\\
\bottomrule[1.3pt]
\end{tabular}
}
\label{table:imagenet-c}
\end{table*}

\vspace{0.2em}
\noindent\textbf{\textit{mini}-DomainNet.} It is constructed from DomainNet~\cite{peng2019moment} dataset, which is for the domain adaptation task originally. It includes six domains, which are \textit{clipart}, \textit{infograph}, \textit{painting}, \textit{quickdraw}, \textit{real} and \textit{sketch}. There are 345 categories of common objects in each domain. Since GFSCL needs to execute the class- and domain-incremental learning, we only use the first 80 categories. We employ 60 and 20 classes in the base session and incremental learning sessions, respectively. Notably, the incremental sessions are conducted with the 5-way 5-shot learning pattern. The learning pattern may be class- or domain-incremental learning. For each incremental session, training data is all from the same domain. There are 9 sessions in total, and for each novel class, the training samples of two domains arrive in two neighboring sessions in sequential order. The unseen domain is \textit{painting}, and the detailed protocols can be found in Table~\ref{table:domainet-protocol}.

\begin{table*}[t]
\caption{Accuracy of Base and Novel classes in \textit{mini}-DomainNet. \textit{Average Forgetting} is to evaluate the forgetting issues of base categories, and \textit{average Acc.} is to assess the overfitting issue of novel categories.}
%Our proposed UaD-CE achieves the state-of-the-art results with respect to the three indicators.
\centering
\setlength{\tabcolsep}{11pt}
\resizebox{15.5cm}{!}{
\tabcolsep 0.06in
\begin{tabular}{cccccccccccccc}
\bottomrule[1.3pt]
\multirow{2}{*}{\textbf{Backbone}} & \multirow{2}{*}{\textbf{Method}} & \multirow{2}{*}{\textbf{Classes}}& \multicolumn{9}{c}{\textbf{Session ID}} & \multirow{2}{*}{\makecell[c]{\textbf{Average} \\ \textbf{Forgetting}}} & \multirow{2}{*}{\makecell[c]{\textbf{Average} \\ \textbf{Acc. (Novel)}}}\\ 
\cline{4-12} 
 & & & \textbf{1} & \textbf{2} &\textbf{3}
 &\textbf{4} & \textbf{5}& \textbf{6}&\textbf{7} & \textbf{8}& \textbf{9}\\
\bottomrule[1.3pt]

\multirow{10}{*}{\textbf{\makecell[c]{\textbf{ViT+} \\ \textbf{Adapter}}}}& \multirow{2}{*}{Finetune} & Base & 89.69 &40.97 & 73.31 &41.06& 53.76& 61.58 & 73.93 &47.10 & 62.05 & 32.97 & -\\
& & Novel & - & 99.60 & 98.62 & 77.83 &85.20 & 26.58 & 52.08 & 3.91 & 22.10 & - & 58.25\\
\cline{2-14} 
& \multirow{2}{*}{EWC~\cite{kirkpatrick2017overcoming}} & Base & 89.18 & 58.90 & 48.20& 35.36& 61.46& 79.55& 80.98& 66.58& 70.39& 26.50& -\\
& & Novel &- & 99.50 & 98.81& 73.95& 87.41&  43.98& 56.06& 4.03& 21.61& - & 60.67\\
\cline{2-14}
& \multirow{2}{*}{LwF~\cite{li2017learning}} & Base & 89.12 &  56.40& 58.13& 27.24& 51.03& 78.78& 79.44& 56.33& 64.92 & 30.08 &- \\
& & Novel & - & 99.09 & 98.26 &68.85& 85.89& 63.81& 67.87& 10.69& 29.65& - & 65.51\\
\cline{2-14} 
&  \multirow{2}{*}{\textbf{CMoA}} & Base &90.42 & 61.46 & 76.56&  47.95& 64.47& 79.69& 80.19& 75.89& \textbf{77.99}& \textbf{19.90}&-\\ 
& & Novel & -& 98.89 & 97.43& 79.77& 87.95 & 63.05& 68.06& 23.31& \textbf{34.08} &- & \textbf{69.07}\\
\bottomrule[1.3pt]
\end{tabular}
}
\label{table:domainet-basenovel}
\end{table*}

\vspace{0.2em}
\noindent\textbf{\textit{mini}-ImageNet-C.} We organize this dataset from ImageNet-C, a robustness test set including images with common corruptions and perturbations such as blurring, fogging, rotation, and scaling. Images from ImageNet-C are split into 15 domains. We organize \textit{mini}-ImageNet-C following the category list of \textit{mini}-ImageNet~\cite{vinyals2016matching}. We use 60 and 40 classes in the base session and incremental learning sessions, respectively. The incremental sessions are conducted in a 10-way 5-shot learning manner. The 40 novel categories arrive in the 12 sessions, and each category belonging to three domains is learned in three consecutive incremental sessions. For the three incremental sessions, the first one is the class-incremental session, and the other two are domain-incremental sessions. The detailed protocols can be found in Table~\ref{table:imagenet-protocol}. 

\vspace{0.2em}
\noindent\textbf{Evaluation Metrics.} The targets of GFSCL are three-fold: (1) Alleviating catastrophic forgetting of previously seen categories and domains. (2) Mitigating the overfitting issue on novel categories or old categories from new domains arrived in incremental sessions. (3) Enhancing the generalization ability in unseen domains. To evaluate the performance of the GFSCL task, we propose the following evaluation metrics: Final overall classification accuracy, the average accuracy of all sessions, average Domain Generalization Enhancement (DGE), final domain generalization ability, average forgetting of the base categories, and average accuracy of novel categories. We assume that the overall classification performance of seen classes from seen domains and unseen domains in session $i$ is $\alpha_i$ and $\delta_i$, respectively, and the total session number is $S$. The average accuracy of all sessions is defined as
\begin{equation}
\bar\alpha =  \frac{\sum_{i=1}^{S} \alpha _i}{S}.
\label{eq:average_acc}
\end{equation}

Average DGE evaluates the generalization enhancement of the model when more training samples of other domains are introduced in domain-incremental sessions. Assume that $i$ is the session ID that the model first sees a group of novel categories and $j$ is the last session that this group of novel categories is introduced into the model. In this way, session $i$ is for the class-incremental purpose and sessions $i+1$ to $j$ are for the domain increment. From session $i$ to session $j$, the DGE is computed as 
\begin{equation}
\hat\delta =  \delta_j - \delta_i.
\label{eq:average_acc}
\end{equation}
For average DGE, it is computed by averaging the DGE of different domain-incremental intervals. Assume that the classification performance of base categories from seen domains in session $i$ is $\alpha^b_i$. Average forgetting of the base categories is defined as 
\begin{equation}
\tilde{\alpha}^b = \frac{1}{S-1} \sum_{i=2}^{S} (\alpha^b_1 - \alpha^b_i).
\label{eq:average_acc}
\end{equation}
The average accuracy of novel categories measures the overfitting issue. Assume that the classification performance of novel categories from seen domains in session $i$ is $\alpha^n_i$. The average accuracy of novel categories is defined as 
\begin{equation}
\bar\alpha^n =  \frac{\sum_{i=2}^{S} \alpha ^n_i}{S-1}.
\label{eq:average_acc}
\end{equation}

\subsection{Implementation Details}
\label{sec:Details}
\noindent\textbf{Model configurations.} We employed ViT-Base (224/16) as the backbone and loaded the ImageNet-21k pretrained weights before training. The MLP layer was expanded for class-incremental purpose. During training, only the MLP layer and the proposed MoA module were updated. As for the three Convpass layers in MoA, they shared the first dimension reduction layer (1 x 1 Conv and GELU) and the final dimension increase layer ((1 x 1 Conv and GELU), which made the ensemble MoA module lightweight. During training, the model was optimized by SGD \cite{robbins1951stochastic} (with lr=0.1 and wd=5e-4). For the classification head, we applied the cosine similarity for the classification task by computing the distance between testing features and prototypes.

\vspace{0.2em}
\noindent\textbf{Training details.}
ViT-Base~\cite{dosovitskiy2020image} with 12 transformer blocks was used as the defaulted architecture. For the Convpass~\cite{jie2022convolutional} finetuning, the original and hidden channels were 768 and 8, respectively. The adapter number $P$ was 3. $\gamma$ in Eq.~\ref{eq:l_cos_loose}, $\zeta$ in Eq.~\ref{eq:total-loss-class} and $\tau$ in Eq.~\ref{eq:metric_loss} were assigned 0.3, 0.8 and 1.0, respectively. The experiments were implemented with Pytorch on one NVIDIA A100 GPU. For the first session, 160 epochs were executed for the two benchmark datasets. We set the initial learning rate as 0.001, which was divided by 10 after 80 and 120 epochs. The model was trained with the training batch size of 64. For the following sessions, the learning rate is 0.0005 for the total 100 epochs.

\subsection{Baselines and Comparative Studies}
\label{sec:SOTA}
Firstly, we provide baselines of GFSCL task on the two newly arranged benchmark datasets with ResNet-18~\cite{he2016deep} and ViT~\cite{dosovitskiy2020image} as the backbone, respectively. \textit{Finetune} means simply finetuning with few training samples of new classes or domains, which is the lower bound of the GFSCL performance. Then, we compare our proposed CMoA with state-of-the-art methods regarding Six indicators. The methods for the comparison are EWC~\cite{kirkpatrick2017overcoming}, LwF~\cite{li2017learning}, iCaRL~\cite{rebuffi2017icarl}. iCaRL is the rehearsal-based continual learning method since the exemplars of previously seen categories need to be stored in an extra memory.

\begin{table*}[t]
\caption{Accuracy of Base and novel classes in \textit{mini}-ImageNet-C. \textit{Average Forgetting} is to evaluate the forgetting issues of base categories.}
%Our proposed UaD-CE achieves the state-of-the-art results with respect to the three indicators.
\centering
\setlength{\tabcolsep}{11pt}
\resizebox{\textwidth}{!}{
\tabcolsep 0.06in
\begin{tabular}{cccccccccccccccccc}
\bottomrule[1.3pt]
\multirow{2}{*}{\textbf{Backbone}} & \multirow{2}{*}{\textbf{Method}}  & \multirow{2}{*}{\textbf{Class}} & \multicolumn{13}{c}{\textbf{Session ID}} & \multirow{2}{*}{\makecell[c]{\textbf{Average} \\ \textbf{Forgetting}}} & \multirow{2}{*}{\makecell[c]{\textbf{Average} \\ \textbf{Acc.(Novel)}}}\\ 
\cline{4-16} 
 & & & \textbf{1} & \textbf{2} &\textbf{3}
 &\textbf{4} & \textbf{5}& \textbf{6}&\textbf{7} & \textbf{8}& \textbf{9} & \textbf{10} & \textbf{11} & \textbf{12} & \textbf{13} \\
\bottomrule[1.3pt]

\multirow{10}{*}{\textbf{\makecell[c]{\textbf{ViT+} \\ \textbf{Adapter}}}}& \multirow{2}{*}{Finetune} & Base & 91.25 & 45.08 & 55.90 & 55.36&40.44 & 49.84 & 55.36 & 58.69 & 57.61 & 57.55 & 55.29 & 60.09 & 61.16 & 36.89 & - \\
& & Novel & - & 71.67 & 78.89 & 80.41 & 36.83 & 52.84 & 59.19 & 29.78 & 36.75 & 42.64 & 30.38 &  37.33 & 42.69 & - & 49.95  \\
\cline{2-18} 
& \multirow{2}{*}{EWC~\cite{kirkpatrick2017overcoming}} &Base & 91.60 & 57.91 & 57.65 & 49.05 & 46.93 & 54.58 & 47.99 & 58.07 & 55.65 & 52.42 & 58.10 & 63.77 & 64.17 & 36.08 & - \\
& & Novel & - & 78.89 & 83.33 & 79.32 & 46.72 & 58.93 & 46.70 & 36.25 & 43.81 & 47.58 & 33.38 & 40.81 & 45.52 & - & 53.44 \\
\cline{2-18}
& \multirow{2}{*}{LwF~\cite{li2017learning}} & Base & 91.82 & 44.69 & 56.11& 54.27 & 28.07 & 40.67 & 49.33 & 39.62 & 50.71 & 57.48 & 58.69 & 63.43 & 64.33 & 41.20 & - \\
& & Novel & - & 74.22 & 81.11 & 78.52 & 27.72 & 42.89 & 52.81 & 14.92 & 26.86 & 36.35 & 32.11 & 38.69 & 43.43 & - & 45.80   \\
\cline{2-18} 
&  \multirow{2}{*}{\makecell[c]{\textbf{CMoA} \\ \textbf{(Ours)}}}  &Base & 91.80 & 53.05 & 61.27 & 48.23 & 51.38 & 57.02 & 60.64 & 71.25 & 69.10 & 68.33 & 65.72 & 67.76 & 67.95 & \textbf{29.99} & - \\ 
& & Novel & - & 78.44 & 83.00 & 81.26 & 49.72 & 59.91 & 64.52 & 43.81 & 48.94 & 52.86 & 36.36 & 42.32 & 47.02 & - & \textbf{57.35}  \\
\bottomrule[1.3pt]
\end{tabular}
}
\label{table:imagenet-basenovel}
\end{table*}

\vspace{0.2em}
\noindent \textbf{Baselines.} We provide the baselines of GFSCL on the two benchmark datasets with ResNet-18 and ViT as the backbone. The results are illustrated in Fig.~\ref{domainnet} for \textit{mini}-DomainNet and Table~\ref{table:imagenet-c} (Rows 3-20) for \textit{mini}-ImageNet-C. With ResNet-18 as the backbone, we load the pre-trained model and freeze the first three layers. From Fig.~\ref{domainnet}, it can be concluded that the overall classification ability or the domain generalization is of poor performance no matter which method we use. This also reflects that the novel GFSCL is a challenging task. With ViT as the backbone, we load the pre-trained ViT base model and freeze all parameters except the MLP layer. The performance achieves improvements to a large extent, while the performance deterioration is still huge. Moreover, the generalization ability is not satisfactory in view of the lower average DGE and final generalization performance in unseen domains.

\begin{figure*}[t]
\includegraphics[width=1.0\linewidth]{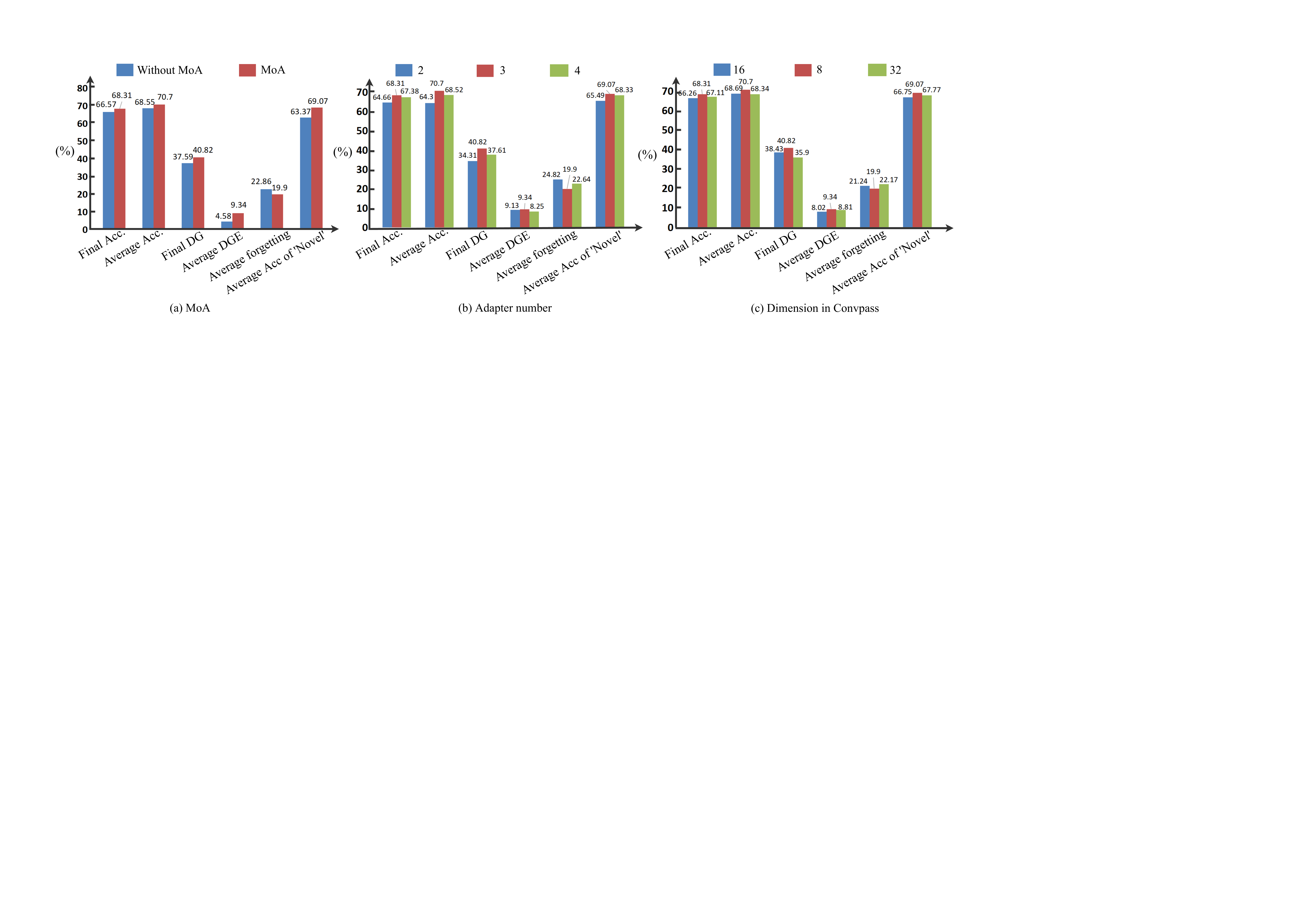}
\centering
\caption{Ablation studies on (a) the mixture of adapters, (b) the number of adapters, and (c) the hidden dimension in Convpass. 'Without MoA' means removing the dynamic weighting module and
cosine similarity regularization.}
\label{fig:ablation-3} 
\end{figure*}

\begin{figure}[t]
\includegraphics[width=1.0\linewidth]{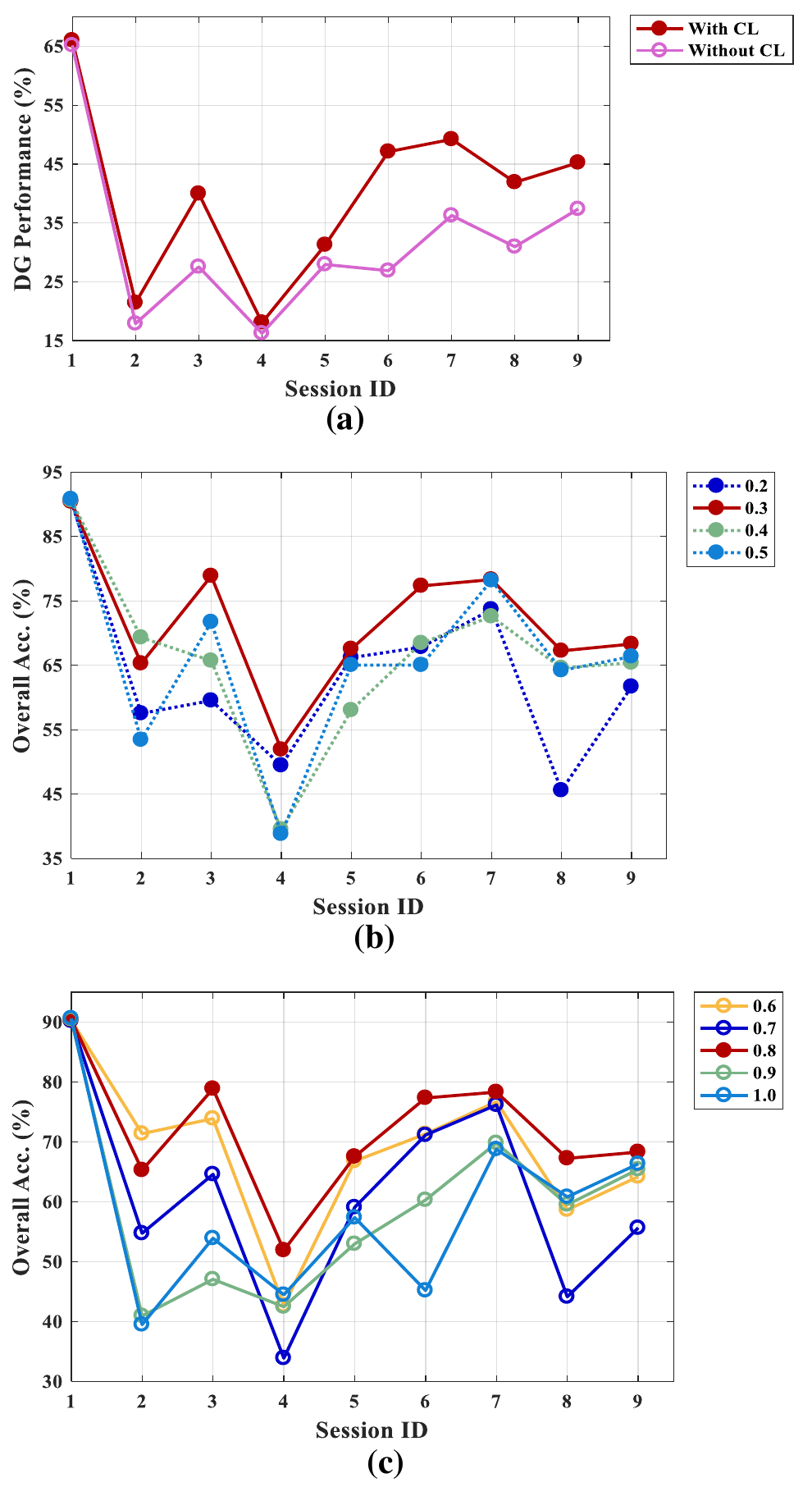}
\centering
\caption{Ablation studies on (a) prototype-calibrated contrastive learning, (b) $\gamma$, and (c) $\zeta$.}
\label{ablation-4-5} 
\end{figure}

\vspace{0.2em}
\noindent\textbf{Results of \textit{mini}-DomainNet.} As shown in Fig.~\ref{domainnet}, we compare our CMoA with the current methods that are all implemented with ViT and adapters as the backbone. As for the overall classification accuracy, it can be seen that the curve representing our achievement locates above all other methods except iCaRL~\cite{rebuffi2017icarl}, which belongs to rehearsal-based methods. While for the final overall classification accuracy of the last session, our method (68.31\%) outperforms iCaRL (67.83\%). This phenomenon also demonstrates that our framework can better tackle the catastrophic forgetting issue. Moreover, as for the indicator of average accuracy, the result obtained by our proposed CMoA (70.70\%) exceeds ViT with iCaRL (62.98\%)and ViT with adapters and iCaRL (62.97\%). For the assessment of the domain generalization performance in unseen domains presented in Fig.~\ref{domainnet} (b), CMoA achieves the surpassing performance of the final classification capability in unseen domains. Furthermore, it is worth noting that the curve representing our method experiences a sharp uptrend in domain-incremental sessions, which represents that our CMoA achieves significant improvement after seeing a new domain. This phenomenon illustrates CMoA can capture domain-invariant representation and further alleviate forgetting previous domains. For the average forgetting of the base categories and average accuracy of novel categories presented in Table~\ref{table:domainet-basenovel}, our CMoA outperforms other methods with less forgetting and higher average accuracy of novel categories. The phenomenon illustrates that our proposed method can also better alleviate the forgetting and overfitting issues.

\begin{table*}[t]
\caption{The results obtained with the mixture of vanilla adapters. \textit{Average Acc.} means the overall classification test on seen classes and domains. \textit{DG} is the generalization ability in unseen domains.}
%Our proposed UaD-CE achieves the state-of-the-art results with respect to the three indicators.
\centering
\setlength{\tabcolsep}{12pt}
\resizebox{11.0cm}{!}{
\tabcolsep 0.06in
\begin{tabular}{cccccccccccccc}
\bottomrule[1.3pt]
\multirow{2}{*}{\textbf{Eval.}}& \multicolumn{9}{c}{\textbf{Session ID}} & \multirow{2}{*}{\makecell[c]{\textbf{Average} \\ \textbf{Acc.}}} &
\multirow{2}{*}{\makecell[c]{\textbf{Average} \\ \textbf{DGE}}} \\ 
\cline{2-10} 
 & \textbf{1} & \textbf{2} &\textbf{3}
 &\textbf{4} & \textbf{5}& \textbf{6}&\textbf{7} & \textbf{8}& \textbf{9}\\
\bottomrule[1.3pt]
Acc. & 89.31 & 76.81 & 77.37 & 47.63 & 66.63 & 73.28 & 75.42 &62.24  & 66.40 & 70.57 &-\\
DG & 65.52 & 36.40 &35.52 & 18.00 & 29.00& 40.78 &42.70 & 30.02& 37.65& -& 4.92\\ 
\bottomrule[1.3pt]
\end{tabular}
}
\label{table:adapter_type}
\end{table*} 

\begin{table*}[t]
\caption{The \textit{Top-5} accuracy of \textit{mini}-DomainNet obtained with different parameter-efficient tuning methods and VPT-based continual learning methods. \textit{Average Acc.} means the overall classification test on seen classes and domains. }
%Our proposed UaD-CE achieves the state-of-the-art results with respect to the three indicators.
\centering
\setlength{\tabcolsep}{12pt}
\resizebox{11.0cm}{!}{
\tabcolsep 0.06in
\begin{tabular}{ccccccccccccc}
\bottomrule[1.3pt]
\multirow{2}{*}{\textbf{Method}}& \multicolumn{9}{c}{\textbf{Session ID}} & \multirow{2}{*}{\makecell[c]{\textbf{Average} \\ \textbf{Acc.}}} \\ 
\cline{2-10} 
 & \textbf{1} & \textbf{2} &\textbf{3}
 &\textbf{4} & \textbf{5}& \textbf{6}&\textbf{7} & \textbf{8}& \textbf{9}\\
\bottomrule[1pt]
ViT+LoRA~\cite{hu2021lora} & 96.70  & 76.51  & 87.25 & 33.51 & 68.01 & 53.43 &72.30 & 30.19 & 57.67 & 63.95 \\
ViT+VPT~\cite{jia2022visual} & 97.50 & 59.01 & 65.39 & 20.55 & 38.96 & 25.29 & 41.22  & 44.93 & 52.23 & 49.45\\ 
\bottomrule[1pt]
L2P~\cite{wang2022learning} & 92.18 & 75.78 & 71.50 & 56.58 & 51.77 & 44.83 & 40.93 & 37.21 &32.88 & 55.96 \\ 
DualPrompt~\cite{wang2022dualprompt} & 92.40 & 80.59 & 69.29 & 61.73 & 50.59 & 44.32 &41.82& 39.17 & 32.01 & 56.88 \\
\bottomrule[1pt]
\textbf{CMoA (Ours)} & 96.65 & 68.93 & 80.75 & 57.01 & 70.25 & 90.11 & 90.34 & 74.34 & \textbf{79.46} & \textbf{78.65} \\
\bottomrule[1.3pt]
\end{tabular}
}
\label{table:tuningmethod}
\end{table*}

\vspace{0.2em}
\noindent\textbf{Results of \textit{mini}-ImageNet-C.} 
We present the comparative study results in Table~\ref{table:imagenet-c} (Rows 21-31) with ViT and adapters as the backbone. CMoA can promote more domain generalization ability when the samples from new domains are imported into the model. Compared with other rehearsal-free methods, our method achieves remarkable performance in the final overall classification accuracy and final generalization ability. However, when compared with the rehearsal-based method, our CMoA can not exceed with regard to the average classification accuracy of all sessions and domain generalization ability. Rehearsal-based methods require storing old samples in an extra memory, which is much easier to memorize previous discriminative capabilities. To further validate that our proposed CMoA can better fulfill the stability-plasticity dilemma, the average forgetting of the base categories and average accuracy of novel categories are presented in Table~\ref{table:imagenet-basenovel}. It can be concluded that our CMoA suffers from less forgetting and achieves the higher average accuracy of novel categories when compared with our rehearsal-free methods. 

\subsection{Ablation Studies}

\vspace{0.2em}
\noindent\textbf{Efficacy of mixture of adapters.}
Here we evaluate the efficacy of MoA, which aims to capture more comprehensive knowledge, and we present comparative results with the configuration of removing the dynamic weighting module and cosine similarity regularization in Fig.~\ref{fig:ablation-3}(a). When we assign the same 3 adapters, our dynamic MoA achieves superior performance regarding 6 evaluation indicators. Therefore, with dynamic MoA, the model suffers from less catastrophic forgetting and generalizes better to new domains. Since the incorporated dynamic MoA is designed to force each adapter to focus on different classes; thus the overfitting issue can be mitigated to some extent and the forgetting issue can deteriorate when the model is modified by new classes.

\begin{figure}[t]
\includegraphics[width=0.9\linewidth]{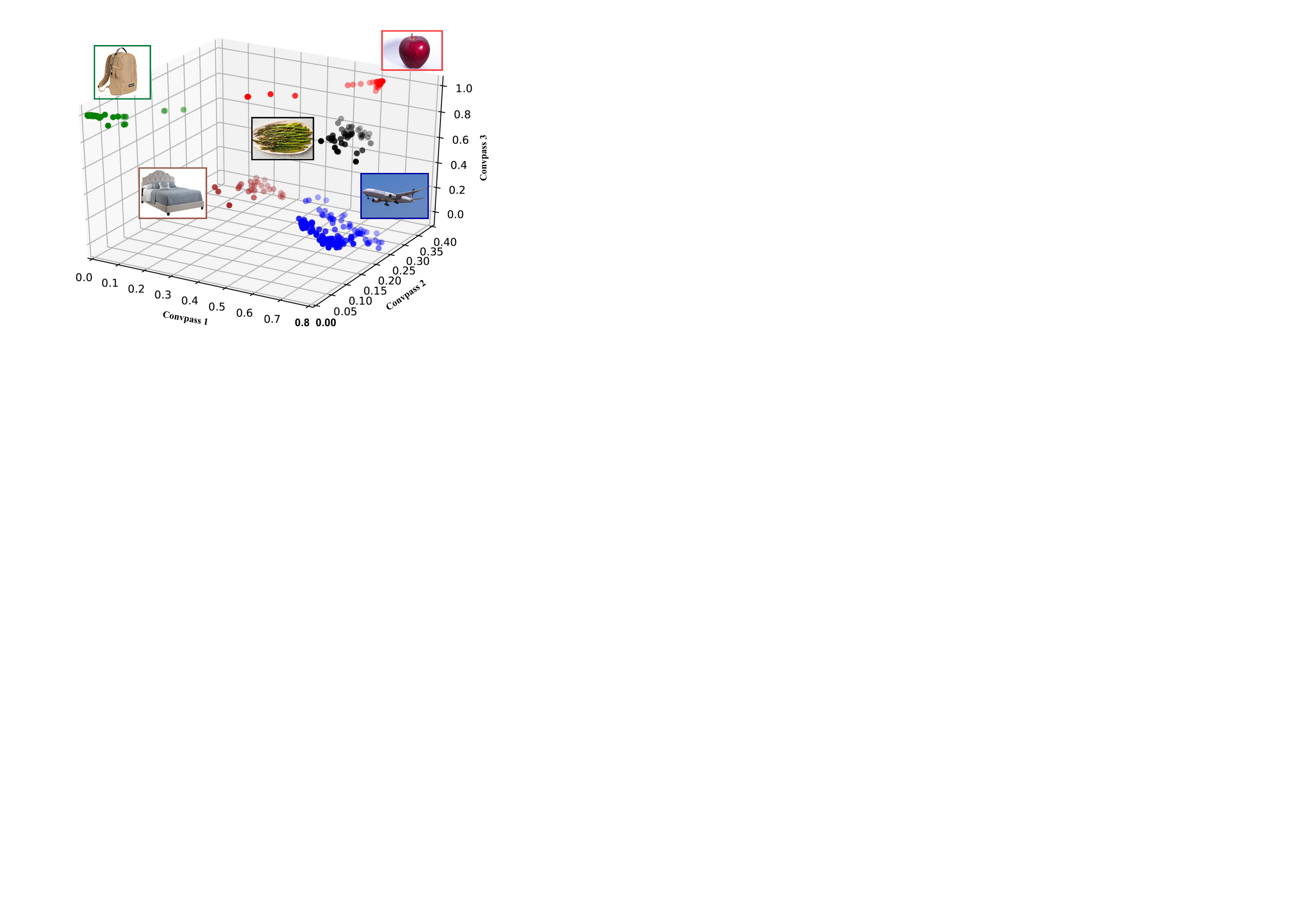}
\centering
\caption{Visualization of dynamic weighting results of adapter experts in the last transformer layer.}
\label{attention-visual} 
\end{figure}

\vspace{0.2em}
\noindent\textbf{Impact of the number of adapters.} 
We present the ablation study results in Fig.~\ref{fig:ablation-3}(b). Towards the evaluation indicators referring to classification accuracy on seen categories and seen domains, the performance deteriorates with 2 or 4 adapters. Since we hope each adapter focuses on different categories, 2 adapters are not enough for 85 categories, which may cause ambiguity among classes. With 4 adapters, there may exist conflicts when we add the cosine similarity regularization on any two adapters. 

\vspace{0.2em}
\noindent\textbf{Impact of the hidden dimension in Convpass.}
The dimension $\beta$ of Convpass decides the extra added number of parameters to ViT framework, and $\beta$ is expected to be smaller. In our experiments, the default $\beta$ is 8, and we compare the obtained results with the dimension of 16 and 32 in Fig.~\ref{fig:ablation-3}(c). When $\beta$ is 8, CMoA obtains outstanding performance in the classification task of both seen and unseen domains. With 16 and 32 dimensions, the performance drops, which may result from that training samples are not enough to fit more learnable parameters. Moreover, the model commonly possesses decent generalization ability with fewer parameters.

%\textcolor{red}{Since novel classes contain limited training data, more learnable parameters are less prone to overfitting. Moreover, the model commonly possesses decent generalization ability with fewer parameters.}

\vspace{0.2em}
\noindent\textbf{Comparison with the vanilla adapter.}
To further validate the efficacy of CMoA, we compare it by replacing the Convpass with the vanilla adapter module~\cite{houlsby2019parameter}, and the results are illustrated in Table~\ref{table:adapter_type}. The classification performance in seen and unseen domains is inferior to that of the configuration with Convpass as the adapter module. It also validates that using Convpass in our CMoA can efficiently alleviate catastrophic forgetting and enhance the generalization ability. This is because of the hard-coded inductive bias of convolutional layers in Convpass~\cite{jie2022convolutional} that is more suitable for visual tasks. Moreover, since ViT is a strong representation framework with a large number of parameters, using vanilla adapters with more added parameters may become harder to handle the overfitting issue.

%Convpass is a bypass layer that introduces only a small amount of trainable parameters to adapt the large ViT, while the vanilla adapter is an extra layer for the large ViT. Hence, there are more parameters when using original adapters. Moreover, t

\vspace{0.2em}
\noindent\textbf{Efficacy of Prototype-Calibrated Contrastive Learning.}
We conduct this ablation study by removing $\mathcal{L}^{c}$ in Eq.~\ref{eq:total-loss-domain} when the domain-incremental is conducted. The results are shown in Fig.~\ref{ablation-4-5}(a). It can be concluded that the incorporated contrastive loss can improve domain-invariant representation learning. Given limited labeled data, the overfitting issue is too severe to accurately model the class distribution. With the proposed contrastive loss, samples belonging to the same category are concentrated in the feature level by decreasing the intra-class variation. In this way, the model can feature the domain-invariant class representation, which is more approximate to the real distribution and affable for the domain-incremental and domain generalization tasks.

\begin{figure*}[t]
\includegraphics[width=0.75\linewidth]{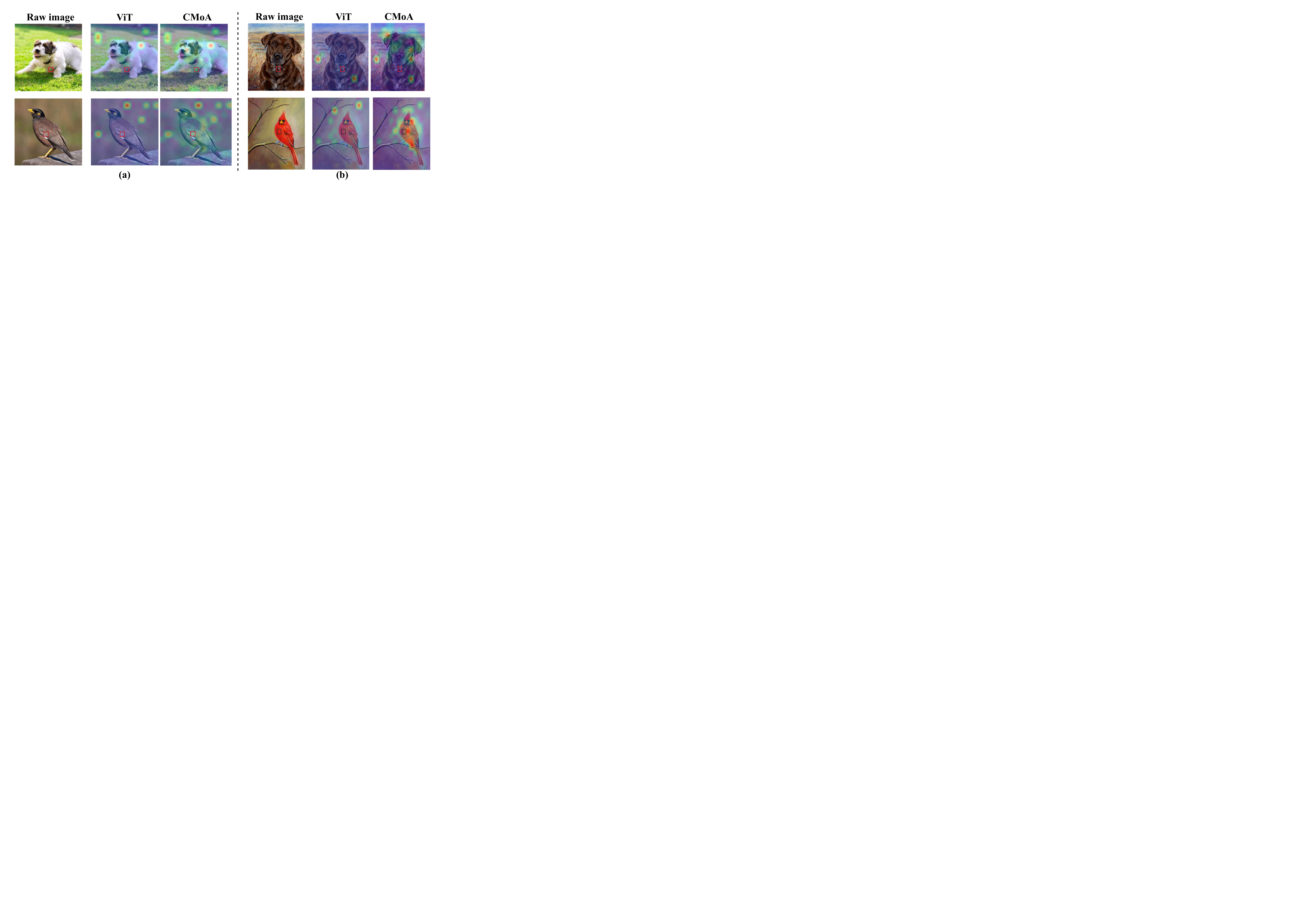}
\centering
\caption{Visualization of attention map of the last transformer layer with samples from (a) seen domains and (b) unseen domains.}
\label{visualization} 
\end{figure*}

\vspace{0.2em}
\noindent\textbf{Impact of $\gamma$ and $\zeta$.}
In these experiments, we evaluate the impacts of loose hyperparameter $\gamma$ in Eq.~\ref{eq:l_cos_loose} and the cosine similarity loss weight $\zeta$ in Eq.~\ref{eq:total-loss-class} by assigning different values. The hyperparameter $\gamma$ and the loss weight $\zeta$ determine the strength of the cosine similarity regularization. By giving different values to $\gamma$ and $\zeta$, the ablation study results are illustrated in Fig.~\ref{ablation-4-5}(b) and (c), and it can be summarized that CMoA achieves the superior discriminative ability to detect seen categories and domains when $\gamma$ and $\zeta$ are assigned 0.3 and 0.8, respectively.

\vspace{0.2em}
\noindent\textbf{Comparison with other parameter-efficient tuning methods.}
Currently, there are mainly three parameter-efficient tuning methods for ViT: adapter~\cite{houlsby2019parameter, jie2022convolutional}, Visual Prompt Tuning (VPT)~\cite{jia2022visual}, and Low-Rank Adaptation (LoRA)~\cite{hu2021lora}. In this paper, we use adapters as the tuning method in our proposed framework. Here we provide the results of \textit{mini}-DomainNet implemented with VPT, LoRA and other two VPT-based continual learning methods (L2P~\cite{wang2022learning} and DualPrompt~\cite{wang2022dualprompt}) in Table~\ref{table:tuningmethod}. It is worth noting that the results are all \textit{Top-5} accuracy. Our CMoA outperforms other parameter-tuning methods and VPT-based continual learning method.

\subsection{Further Remarks and Discussions}
\noindent\textbf{Visualization of dynamic weights.} To validate that each adapter in the MoA module mainly focuses on different categories, we visualize the dynamic weighting results of the last transformer layer in Fig.~\ref{attention-visual}. It can be seen that the dynamic weighting results show discriminative distributions and a specific adapter is assigned the larger weight for each category. For example, the first adapter mainly concentrates on ``Airplane'' and the third adapter works more on ``Backpack'' and ``apple''. This statistical phenomenon validates the efficacy of the dynamic weighting operation, and it is also benefited from our CMoA with the cosine similarity constraint firstly to capture different information and then the prototype-calibrated contrastive learning to obtain a discriminative distribution.

\vspace{0.2em}
\noindent\textbf{Visualization of attention maps.} We visualize the attention maps of the last transformer layer with the samples from seen domains in Fig.~\ref{visualization}(a) and unseen domains in Fig.~\ref{visualization}(b). For these two groups, the first image is the raw data, and the second image and the third image represent the attention map obtained by ordinary ViT and our CMoA, respectively. Given a patch framed by the red or black rectangle, our CMoA can possess distinguishable attention to related areas than that of ViT. It can be concluded that our proposed method can capture more accurate long-range dependencies when given a small patch framed by the red or black rectangle in the figure. This phenomenon results from that the mixture of adapter is good at learning specific knowledge, which can guide the whole framework to focus on salient areas.

\begin{table}[t]
\caption{Another protocol of \textit{mini}-DomainNet. Here we only discuss the overall classification accuracy.}
%Our proposed UaD-CE achieves the state-of-the-art results with respect to the three indicators.
\centering
\setlength{\tabcolsep}{6pt}
\resizebox{0.48\textwidth}{!}{
\tabcolsep 0.06in
\begin{tabular}{c|cccccc}
\bottomrule[1.3pt]
{\diagbox{\textbf{Item}}{\makecell[c]{\textbf{Session} \\ \textbf{ID}}}}& \textbf{1}  & \textbf{2} &\textbf{3}
 &\textbf{4} & \textbf{5}& \textbf{6}\\
\bottomrule[1.3pt]
Task & CIL, DIL & CIL, DIL & CIL, DIL & CIL, DIL & CIL, DIL & CIL, DIL \\
\cline{1-7} 
 Domain & Real & Painting & Clipart & Sketch  & Infograph & Quickdraw  \\
\cline{1-7} 
Class ID & 0-59 & 60-64 & 65-69  & 70-74 & 75-79 & 80-84 \\
\cline{1-7} 
Dataset & Large-scale & \makecell[c]{5-Way \\ 5-Shot} & \makecell[c]{5-Way \\ 5-Shot} & \makecell[c]{5-Way \\ 5-Shot} & \makecell[c]{5-Way \\ 5-Shot} & \makecell[c]{5-Way \\ 5-Shot}  \\
\bottomrule[1.3pt]
\end{tabular}
}
\label{table:domainet-protocol-2}
\end{table}

\begin{figure}[t]
\includegraphics[width=0.85\linewidth]{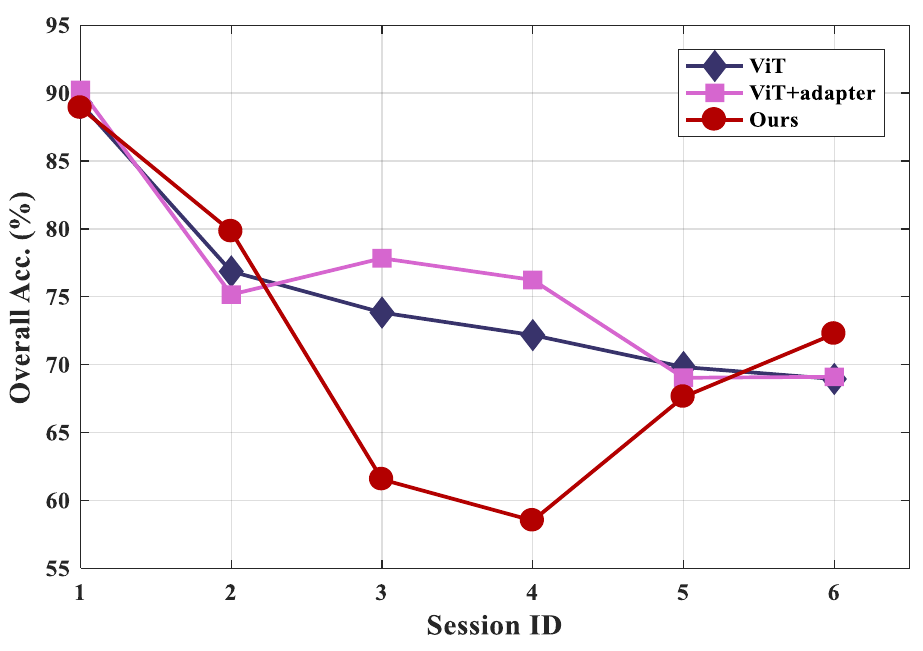}
\centering
\caption{Overall classification performance of \textit{mini}-DomainNet when new classes from new domain arrive consequently.}
\label{figure:discussion} 
\end{figure}

\vspace{0.2em}
\noindent\textbf{Discussions about the protocol.}
For the configuration of GFSCL, new classes or old classes from new domains are encountered for each incremental session. Here we discuss another scenario that new classes from new domain arrives in each session. It is more challenging than the current configuration, since there is no salient previous knowledge that can be transferred to assist the current learning procedure. We use \textit{mini}-DomainNet as an example to explain this configuration and present the classification performance. For the base session, we employ 60 classes from \textit{Real} domain. The incremental sessions are also conducted with the 5-way 5-shot learning pattern. The learning pattern is class- and domain-incremental learning, thus there are totally 6 sessions. The detailed protocol is presented in Table~\ref{table:domainet-protocol-2}. Due to this configuration, there are no prototypes from previous sessions to guide current classes. In this way, we implement the proposed CMoA with common contrastive learning here by removing the prototypes from the sampling set. The results are illustrated in Fig.~\ref{figure:discussion}. Though the final overall classification performance achieved by our method is surpassing, our method suffers from severe performance deterioration in the first 4 sessions and even is inferior to clean ViT and ViT with adapters. In the future work, we will dedicate to explore solutions to tackle this issue of this more practical and challenging scenario.

\section{Conclusion}
In this paper, we proposed an unexplored configuration named General Few-Shot Continual Learning (GFSCL) which aims to cumulatively learn novel classes or domains and evaluate the generation ability in unseen domains contemporarily. For GFSCL, we arranged two benchmark datasets and built baselines based on three backbones. Furthermore, we raised the Contrastive Mixture of Adapters (CMoA) based on ViT containing the dynamic mixture of adapters for the class-incremental purpose and prototype-calibrated contrastive learning for the domain-related issues. The exhaustive evaluations of the proposed configuration and modules on two benchmark datasets are conducted to certify the efficacy regarding six evaluation indicators. In the future, it is potential to investigate more practical and challenging GFSCL configurations, such as new classes from new domains arriving in each session. 
Besides the protocols, we will explore more generic and efficient GFSCL frameworks that can adapt to different task configurations. 

\vspace{0.4em}
\noindent\textbf{Acknowledgment}  \quad This work was partially supported by National Key Research and Development Program of China No. 2021YFB3100800, the Academy of Finland under grant 331883, and the China Scholarship Council (CSC) under grant 201903170129.

\section{Data Availability Statement}
The data that support the findings of this study are openly available in the DomainNet\footnote{http://ai.bu.edu/DomainNet/\#dataset} and ImageNet-C\footnote{ https://github.com/hendrycks/robustness/}. All the codes are publicly available at \href{https://github.com/yawencui/CMoA}{https://github.com/yawencui/CMoA}.

% 7376257 params   7.03 M
% 50481262240 FLOPs  47.01 G
% inputs_temp = torch.randn(1, 3, 160, 128, 128)
%  flops, params = profile(model_2DViT, 
% inputs=(inputs_temp, gra_sharp))

%\noindent\textbf{Acknowledgment}  \quad This work was supported by the Academy of Finland (Academy Professor project EmotionAI with grant numbers 336116 and 345122, and ICT2023 project with grant number 345948), the National Natural Science Foundation of China (Grant No. 62002283, the EPSRC grant: Turing AI Fellowship: EP/W002981/1, EPSRC/MURI grant EP/N019474/1. We would also like to thank the Royal Academy of Engineering and FiveAI. As well, the authors wish to acknowledge CSC-IT Center for Science, Finland, for computational resources.

%\bibliographystyle{bst/sn-apacite}
\bibliographystyle{apalike}
\bibliography{sn-bibliography}% common bib file

\begin{thebibliography}{}

\bibitem[Ahmad et~al., 2022]{ahmad2022few}
Ahmad, T., Dhamija, A.~R., Cruz, S., Rabinowitz, R., Li, C., Jafarzadeh, M.,
  and Boult, T.~E. (2022).
\newblock Few-shot class incremental learning leveraging self-supervised
  features.
\newblock In {\em Proceedings of the IEEE/CVF Conference on Computer Vision and
  Pattern Recognition}, pages 3900--3910.

\bibitem[Aljundi et~al., 2018]{aljundi2018memory}
Aljundi, R., Babiloni, F., Elhoseiny, M., Rohrbach, M., and Tuytelaars, T.
  (2018).
\newblock Memory aware synapses: Learning what (not) to forget.
\newblock In {\em Proceedings of the European Conference on Computer Vision
  (ECCV)}, pages 139--154.

\bibitem[Bansal et~al., 2022]{bansal2022meta}
Bansal, T., Alzubi, S., Wang, T., Lee, J.-Y., and McCallum, A. (2022).
\newblock Meta-adapters: Parameter efficient few-shot fine-tuning through
  meta-learning.
\newblock In {\em International Conference on Automated Machine Learning},
  pages 19--1. PMLR.

\bibitem[Bateni et~al., 2020]{bateni2020improved}
Bateni, P., Goyal, R., Masrani, V., Wood, F., and Sigal, L. (2020).
\newblock Improved few-shot visual classification.
\newblock In {\em Proceedings of the IEEE/CVF Conference on Computer Vision and
  Pattern Recognition}, pages 14493--14502.

\bibitem[Carlucci et~al., 2019]{carlucci2019domain}
Carlucci, F.~M., D'Innocente, A., Bucci, S., Caputo, B., and Tommasi, T.
  (2019).
\newblock Domain generalization by solving jigsaw puzzles.
\newblock In {\em Proceedings of the IEEE/CVF Conference on Computer Vision and
  Pattern Recognition}, pages 2229--2238.

\bibitem[Cheraghian et~al., 2021a]{cheraghian2021semantic}
Cheraghian, A., Rahman, S., Fang, P., Roy, S.~K., Petersson, L., and Harandi,
  M. (2021a).
\newblock Semantic-aware knowledge distillation for few-shot class-incremental
  learning.
\newblock In {\em CVPR}, pages 2534--2543.

\bibitem[Cheraghian et~al., 2021b]{cheraghian2021synthesized}
Cheraghian, A., Rahman, S., Ramasinghe, S., Fang, P., Simon, C., Petersson, L.,
  and Harandi, M. (2021b).
\newblock Synthesized feature based few-shot class-incremental learning on a
  mixture of subspaces.
\newblock In {\em ICCV}, pages 8661--8670.

\bibitem[Chi et~al., 2022]{chi2022metafscil}
Chi, Z., Gu, L., Liu, H., Wang, Y., Yu, Y., and Tang, J. (2022).
\newblock Metafscil: A meta-learning approach for few-shot class incremental
  learning.
\newblock In {\em Proceedings of the IEEE/CVF Conference on Computer Vision and
  Pattern Recognition}, pages 14166--14175.

\bibitem[De~Lange et~al., 2021]{de2021continual}
De~Lange, M., Aljundi, R., Masana, M., Parisot, S., Jia, X., Leonardis, A.,
  Slabaugh, G., and Tuytelaars, T. (2021).
\newblock A continual learning survey: Defying forgetting in classification
  tasks.
\newblock {\em IEEE transactions on pattern analysis and machine intelligence},
  44(7):3366--3385.

\bibitem[Doersch et~al., 2020]{doersch2020crosstransformers}
Doersch, C., Gupta, A., and Zisserman, A. (2020).
\newblock Crosstransformers: spatially-aware few-shot transfer.
\newblock {\em Advances in Neural Information Processing Systems},
  33:21981--21993.

\bibitem[Dong et~al., 2021]{dong2021few}
Dong, S., Hong, X., Tao, X., Chang, X., Wei, X., and Gong, Y. (2021).
\newblock Few-shot class-incremental learning via relation knowledge
  distillation.
\newblock In {\em AAAI}, pages 1255--1263.

\bibitem[Dosovitskiy et~al., 2020]{dosovitskiy2020image}
Dosovitskiy, A., Beyer, L., Kolesnikov, A., Weissenborn, D., Zhai, X.,
  Unterthiner, T., Dehghani, M., Minderer, M., Heigold, G., Gelly, S., et~al.
  (2020).
\newblock An image is worth 16x16 words: Transformers for image recognition at
  scale.
\newblock In {\em International Conference on Learning Representations}.

\bibitem[Douillard et~al., 2022]{douillard2022dytox}
Douillard, A., Ram{\'e}, A., Couairon, G., and Cord, M. (2022).
\newblock Dytox: Transformers for continual learning with dynamic token
  expansion.
\newblock In {\em Proceedings of the IEEE/CVF Conference on Computer Vision and
  Pattern Recognition}, pages 9285--9295.

\bibitem[Fini et~al., 2022]{fini2022self}
Fini, E., da~Costa, V. G.~T., Alameda-Pineda, X., Ricci, E., Alahari, K., and
  Mairal, J. (2022).
\newblock Self-supervised models are continual learners.
\newblock In {\em Proceedings of the IEEE/CVF Conference on Computer Vision and
  Pattern Recognition}, pages 9621--9630.

\bibitem[Finn et~al., 2017]{finn2017model}
Finn, C., Abbeel, P., and Levine, S. (2017).
\newblock Model-agnostic meta-learning for fast adaptation of deep networks.
\newblock In {\em ICML}, pages 1126--1135.

\bibitem[He et~al., 2016]{he2016deep}
He, K., Zhang, X., Ren, S., and Sun, J. (2016).
\newblock Deep residual learning for image recognition.
\newblock In {\em CVPR}, pages 770--778.

\bibitem[Hendrycks and Dietterich, 2019]{hendrycks2018benchmarking}
Hendrycks, D. and Dietterich, T. (2019).
\newblock Benchmarking neural network robustness to common corruptions and
  perturbations.
\newblock In {\em International Conference on Learning Representations}.

\bibitem[Hersche et~al., 2022]{hersche2022constrained}
Hersche, M., Karunaratne, G., Cherubini, G., Benini, L., Sebastian, A., and
  Rahimi, A. (2022).
\newblock Constrained few-shot class-incremental learning.
\newblock In {\em Proceedings of the IEEE/CVF Conference on Computer Vision and
  Pattern Recognition}, pages 9057--9067.

\bibitem[Hou et~al., 2019]{hou2019learning}
Hou, S., Pan, X., Loy, C.~C., Wang, Z., and Lin, D. (2019).
\newblock Learning a unified classifier incrementally via rebalancing.
\newblock In {\em Proceedings of the IEEE/CVF Conference on Computer Vision and
  Pattern Recognition}, pages 831--839.

\bibitem[Houlsby et~al., 2019]{houlsby2019parameter}
Houlsby, N., Giurgiu, A., Jastrzebski, S., Morrone, B., De~Laroussilhe, Q.,
  Gesmundo, A., Attariyan, M., and Gelly, S. (2019).
\newblock Parameter-efficient transfer learning for nlp.
\newblock In {\em International Conference on Machine Learning}, pages
  2790--2799. PMLR.

\bibitem[Hu et~al., 2021]{hu2021lora}
Hu, E.~J., Shen, Y., Wallis, P., Allen-Zhu, Z., Li, Y., Wang, S., Wang, L., and
  Chen, W. (2021).
\newblock Lora: Low-rank adaptation of large language models.
\newblock {\em arXiv preprint arXiv:2106.09685}.

\bibitem[Hu et~al., 2019]{hu2019overcoming}
Hu, W., Lin, Z., Liu, B., Tao, C., Tao, Z.~T., Zhao, D., Ma, J., and Yan, R.
  (2019).
\newblock Overcoming catastrophic forgetting for continual learning via model
  adaptation.
\newblock In {\em International conference on learning representations}.

\bibitem[Huang et~al., 2020]{huang2020self}
Huang, Z., Wang, H., Xing, E.~P., and Huang, D. (2020).
\newblock Self-challenging improves cross-domain generalization.
\newblock In {\em Computer Vision--ECCV 2020: 16th European Conference,
  Glasgow, UK, August 23--28, 2020, Proceedings, Part II 16}, pages 124--140.
  Springer.

\bibitem[Jia et~al., 2022]{jia2022visual}
Jia, M., Tang, L., Chen, B.-C., Cardie, C., Belongie, S., Hariharan, B., and
  Lim, S.-N. (2022).
\newblock Visual prompt tuning.
\newblock In {\em Computer Vision--ECCV 2022: 17th European Conference, Tel
  Aviv, Israel, October 23--27, 2022, Proceedings, Part XXXIII}, pages
  709--727. Springer.

\bibitem[Jie and Deng, 2022]{jie2022convolutional}
Jie, S. and Deng, Z.-H. (2022).
\newblock Convolutional bypasses are better vision transformer adapters.
\newblock {\em arXiv preprint arXiv:2207.07039}.

\bibitem[Khan and Dai, 2021]{khan2021video}
Khan, S.~A. and Dai, H. (2021).
\newblock Video transformer for deepfake detection with incremental learning.
\newblock In {\em Proceedings of the 29th ACM International Conference on
  Multimedia}, pages 1821--1828.

\bibitem[Kim et~al., 2021]{kim2021cored}
Kim, M., Tariq, S., and Woo, S.~S. (2021).
\newblock Cored: Generalizing fake media detection with continual
  representation using distillation.
\newblock In {\em Proceedings of the 29th ACM International Conference on
  Multimedia}, pages 337--346.

\bibitem[Kirkpatrick et~al., 2017]{kirkpatrick2017overcoming}
Kirkpatrick, J., Pascanu, R., Rabinowitz, N., Veness, J., Desjardins, G., Rusu,
  A.~A., Milan, K., Quan, J., Ramalho, T., Grabska-Barwinska, A., et~al.
  (2017).
\newblock Overcoming catastrophic forgetting in neural networks.
\newblock {\em Proceedings of the national academy of sciences},
  114(13):3521--3526.

\bibitem[Lee et~al., 2019]{lee2019meta}
Lee, K., Maji, S., Ravichandran, A., and Soatto, S. (2019).
\newblock Meta-learning with differentiable convex optimization.
\newblock In {\em Proceedings of the IEEE/CVF conference on computer vision and
  pattern recognition}, pages 10657--10665.

\bibitem[Li et~al., 2019]{li2019episodic}
Li, D., Zhang, J., Yang, Y., Liu, C., Song, Y.-Z., and Hospedales, T.~M.
  (2019).
\newblock Episodic training for domain generalization.
\newblock In {\em Proceedings of the IEEE/CVF International Conference on
  Computer Vision}, pages 1446--1455.

\bibitem[Li et~al., 2022]{li2022cross}
Li, W.-H., Liu, X., and Bilen, H. (2022).
\newblock Cross-domain few-shot learning with task-specific adapters.
\newblock In {\em Proceedings of the IEEE/CVF Conference on Computer Vision and
  Pattern Recognition}, pages 7161--7170.

\bibitem[Li and Hoiem, 2017]{li2017learning}
Li, Z. and Hoiem, D. (2017).
\newblock Learning without forgetting.
\newblock {\em IEEE transactions on pattern analysis and machine intelligence},
  40(12):2935--2947.

\bibitem[Marra et~al., 2019]{marra2019incremental}
Marra, F., Saltori, C., Boato, G., and Verdoliva, L. (2019).
\newblock Incremental learning for the detection and classification of
  gan-generated images.
\newblock In {\em 2019 IEEE international workshop on information forensics and
  security (WIFS)}, pages 1--6. IEEE.

\bibitem[Mirza et~al., 2022]{mirza2022efficient}
Mirza, M.~J., Masana, M., Possegger, H., and Bischof, H. (2022).
\newblock An efficient domain-incremental learning approach to drive in all
  weather conditions.
\newblock In {\em Proceedings of the IEEE/CVF Conference on Computer Vision and
  Pattern Recognition}, pages 3001--3011.

\bibitem[Munkhdalai and Yu, 2017]{munkhdalai2017meta}
Munkhdalai, T. and Yu, H. (2017).
\newblock Meta networks.
\newblock In {\em International Conference on Machine Learning}, pages
  2554--2563. PMLR.

\bibitem[Peng et~al., 2022]{peng2022few}
Peng, C., Zhao, K., Wang, T., Li, M., and Lovell, B.~C. (2022).
\newblock Few-shot class-incremental learning from an open-set perspective.
\newblock In {\em European Conference on Computer Vision}, pages 382--397.
  Springer.

\bibitem[Peng et~al., 2019]{peng2019moment}
Peng, X., Bai, Q., Xia, X., Huang, Z., Saenko, K., and Wang, B. (2019).
\newblock Moment matching for multi-source domain adaptation.
\newblock In {\em Proceedings of the IEEE/CVF international conference on
  computer vision}, pages 1406--1415.

\bibitem[Perez et~al., 2021]{perez2021true}
Perez, E., Kiela, D., and Cho, K. (2021).
\newblock True few-shot learning with language models.
\newblock {\em Advances in Neural Information Processing Systems},
  34:11054--11070.

\bibitem[Piratla et~al., 2020]{piratla2020efficient}
Piratla, V., Netrapalli, P., and Sarawagi, S. (2020).
\newblock Efficient domain generalization via common-specific low-rank
  decomposition.
\newblock In {\em International Conference on Machine Learning}, pages
  7728--7738. PMLR.

\bibitem[Rebuffi et~al., 2017]{rebuffi2017icarl}
Rebuffi, S.-A., Kolesnikov, A., Sperl, G., and Lampert, C.~H. (2017).
\newblock icarl: Incremental classifier and representation learning.
\newblock In {\em Proceedings of the IEEE conference on Computer Vision and
  Pattern Recognition}, pages 2001--2010.

\bibitem[Riemer et~al., 2018]{riemer2018learning}
Riemer, M., Cases, I., Ajemian, R., Liu, M., Rish, I., Tu, Y., and Tesauro, G.
  (2018).
\newblock Learning to learn without forgetting by maximizing transfer and
  minimizing interference.
\newblock In {\em International Conference on Learning Representations}.

\bibitem[Robbins and Monro, 1951]{robbins1951stochastic}
Robbins, H. and Monro, S. (1951).
\newblock A stochastic approximation method.
\newblock {\em The annals of mathematical statistics}, pages 400--407.

\bibitem[Shi et~al., 2021]{shi2021overcoming}
Shi, G., Chen, J., Zhang, W., Zhan, L.-M., and Wu, X.-M. (2021).
\newblock Overcoming catastrophic forgetting in incremental few-shot learning
  by finding flat minima.
\newblock {\em Advances in Neural Information Processing Systems},
  34:6747--6761.

\bibitem[Shon et~al., 2022]{shon2022dlcft}
Shon, H., Lee, J., Kim, S.~H., and Kim, J. (2022).
\newblock Dlcft: Deep linear continual fine-tuning for general incremental
  learning.
\newblock In {\em European Conference on Computer Vision}, pages 513--529.
  Springer.

\bibitem[Simon et~al., 2022]{simon2022generalizing}
Simon, C., Faraki, M., Tsai, Y.-H., Yu, X., Schulter, S., Suh, Y., Harandi, M.,
  and Chandraker, M. (2022).
\newblock On generalizing beyond domains in cross-domain continual learning.
\newblock In {\em Proceedings of the IEEE/CVF Conference on Computer Vision and
  Pattern Recognition}, pages 9265--9274.

\bibitem[Snell et~al., 2017]{snell2017prototypical}
Snell, J., Swersky, K., and Zemel, R. (2017).
\newblock Prototypical networks for few-shot learning.
\newblock {\em Advances in neural information processing systems}, 30.

\bibitem[Sun et~al., 2019]{sun2019meta}
Sun, Q., Liu, Y., Chua, T.-S., and Schiele, B. (2019).
\newblock Meta-transfer learning for few-shot learning.
\newblock In {\em Proceedings of the IEEE/CVF Conference on Computer Vision and
  Pattern Recognition}, pages 403--412.

\bibitem[Tang and Matteson, 2020]{tang2020graph}
Tang, B. and Matteson, D.~S. (2020).
\newblock Graph-based continual learning.
\newblock In {\em International Conference on Learning Representations}.

\bibitem[Tao et~al., 2020a]{tao2020topology}
Tao, X., Chang, X., Hong, X., Wei, X., and Gong, Y. (2020a).
\newblock Topology-preserving class-incremental learning.
\newblock In {\em European Conference on Computer Vision}, pages 254--270.
  Springer.

\bibitem[Tao et~al., 2020b]{tao2020few}
Tao, X., Hong, X., Chang, X., Dong, S., Wei, X., and Gong, Y. (2020b).
\newblock Few-shot class-incremental learning.
\newblock In {\em CVPR}, pages 12183--12192.

\bibitem[Vinyals et~al., 2016]{vinyals2016matching}
Vinyals, O., Blundell, C., Lillicrap, T., Wierstra, D., et~al. (2016).
\newblock Matching networks for one shot learning.
\newblock In {\em NeurPIS}, pages 3630--3638.

\bibitem[Wang et~al., 2022a]{wang2022s}
Wang, Y., Huang, Z., and Hong, X. (2022a).
\newblock S-prompts learning with pre-trained transformers: An occam's razor
  for domain incremental learning.
\newblock {\em arXiv preprint arXiv:2207.12819}.

\bibitem[Wang et~al., 2020a]{wang2020instance}
Wang, Y., Xu, C., Liu, C., Zhang, L., and Fu, Y. (2020a).
\newblock Instance credibility inference for few-shot learning.
\newblock In {\em Proceedings of the IEEE/CVF conference on computer vision and
  pattern recognition}, pages 12836--12845.

\bibitem[Wang et~al., 2020b]{wang2020generalizing}
Wang, Y., Yao, Q., Kwok, J.~T., and Ni, L.~M. (2020b).
\newblock Generalizing from a few examples: A survey on few-shot learning.
\newblock {\em ACM computing surveys (csur)}, 53(3):1--34.

\bibitem[Wang et~al., 2022b]{wang2022dualprompt}
Wang, Z., Zhang, Z., Ebrahimi, S., Sun, R., Zhang, H., Lee, C.-Y., Ren, X., Su,
  G., Perot, V., Dy, J., et~al. (2022b).
\newblock Dualprompt: Complementary prompting for rehearsal-free continual
  learning.
\newblock In {\em Computer Vision--ECCV 2022: 17th European Conference, Tel
  Aviv, Israel, October 23--27, 2022, Proceedings, Part XXVI}, pages 631--648.
  Springer.

\bibitem[Wang et~al., 2022c]{wang2022learning}
Wang, Z., Zhang, Z., Lee, C.-Y., Zhang, H., Sun, R., Ren, X., Su, G., Perot,
  V., Dy, J., and Pfister, T. (2022c).
\newblock Learning to prompt for continual learning.
\newblock In {\em Proceedings of the IEEE/CVF Conference on Computer Vision and
  Pattern Recognition}, pages 139--149.

\bibitem[Wu et~al., 2019]{wu2019large}
Wu, Y., Chen, Y., Wang, L., Ye, Y., Liu, Z., Guo, Y., and Fu, Y. (2019).
\newblock Large scale incremental learning.
\newblock In {\em Proceedings of the IEEE/CVF Conference on Computer Vision and
  Pattern Recognition}, pages 374--382.

\bibitem[Xie et~al., 2022]{xie2022general}
Xie, J., Yan, S., and He, X. (2022).
\newblock General incremental learning with domain-aware categorical
  representations.
\newblock In {\em Proceedings of the IEEE/CVF Conference on Computer Vision and
  Pattern Recognition}, pages 14351--14360.

\bibitem[Xu et~al., 2021]{xu2021fourier}
Xu, Q., Zhang, R., Zhang, Y., Wang, Y., and Tian, Q. (2021).
\newblock A fourier-based framework for domain generalization.
\newblock In {\em Proceedings of the IEEE/CVF Conference on Computer Vision and
  Pattern Recognition}, pages 14383--14392.

\bibitem[Xue et~al., 2022]{xue2022meta}
Xue, M., Zhang, H., Song, J., and Song, M. (2022).
\newblock Meta-attention for vit-backed continual learning.
\newblock In {\em Proceedings of the IEEE/CVF Conference on Computer Vision and
  Pattern Recognition}, pages 150--159.

\bibitem[Yan et~al., 2022]{yan2022learning}
Yan, Q., Gong, D., Liu, Y., van~den Hengel, A., and Shi, J.~Q. (2022).
\newblock Learning bayesian sparse networks with full experience replay for
  continual learning.
\newblock In {\em Proceedings of the IEEE/CVF Conference on Computer Vision and
  Pattern Recognition}, pages 109--118.

\bibitem[Zhang et~al., 2020]{zhang2020deepemd}
Zhang, C., Cai, Y., Lin, G., and Shen, C. (2020).
\newblock Deepemd: Few-shot image classification with differentiable earth
  mover's distance and structured classifiers.
\newblock In {\em Proceedings of the IEEE/CVF conference on computer vision and
  pattern recognition}, pages 12203--12213.

\bibitem[Zhang et~al., 2021]{zhang2021few}
Zhang, C., Song, N., Lin, G., Zheng, Y., Pan, P., and Xu, Y. (2021).
\newblock Few-shot incremental learning with continually evolved classifiers.
\newblock In {\em CVPR}, pages 12455--12464.

\bibitem[Zhang et~al., 2022]{zhang2022tip}
Zhang, R., Zhang, W., Fang, R., Gao, P., Li, K., Dai, J., Qiao, Y., and Li, H.
  (2022).
\newblock Tip-adapter: Training-free adaption of clip for few-shot
  classification.
\newblock In {\em Computer Vision--ECCV 2022: 17th European Conference, Tel
  Aviv, Israel, October 23--27, 2022, Proceedings, Part XXXV}, pages 493--510.
  Springer.

\bibitem[Zhou et~al., 2022a]{zhou2022forward}
Zhou, D.-W., Wang, F.-Y., Ye, H.-J., Ma, L., Pu, S., and Zhan, D.-C. (2022a).
\newblock Forward compatible few-shot class-incremental learning.
\newblock In {\em Proceedings of the IEEE/CVF Conference on Computer Vision and
  Pattern Recognition}, pages 9046--9056.

\bibitem[Zhou et~al., 2022b]{zhou2022few}
Zhou, D.-W., Ye, H.-J., Ma, L., Xie, D., Pu, S., and Zhan, D.-C. (2022b).
\newblock Few-shot class-incremental learning by sampling multi-phase tasks.
\newblock {\em IEEE Transactions on Pattern Analysis and Machine Intelligence}.

\bibitem[Zhou et~al., 2021]{zhou2021domain}
Zhou, K., Yang, Y., Qiao, Y., and Xiang, T. (2021).
\newblock Domain generalization with mixstyle.
\newblock {\em arXiv preprint arXiv:2104.02008}.

\bibitem[Zhu et~al., 2021]{zhu2021self}
Zhu, K., Cao, Y., Zhai, W., Cheng, J., and Zha, Z.-J. (2021).
\newblock Self-promoted prototype refinement for few-shot class-incremental
  learning.
\newblock In {\em Proceedings of the IEEE/CVF Conference on Computer Vision and
  Pattern Recognition}, pages 6801--6810.

\end{thebibliography}



\end{document}